\documentclass[10pt,twocolumn,letterpaper]{article}
\usepackage{xr}
\makeatletter
\newcommand*{\addFileDependency}[1]{
  \typeout{(#1)}
  \@addtofilelist{#1}
  \IfFileExists{#1}{}{\typeout{No file #1.}}
}
\makeatother
\newcommand*{\myexternaldocument}[1]{
    \externaldocument{#1}
    \addFileDependency{#1.tex}
    \addFileDependency{#1.aux}
}
\newcommand\blfootnote[1]{%
  \begingroup
  \renewcommand\thefootnote{}\footnote{#1}%
  \addtocounter{footnote}{-1}%
  \endgroup
}

\myexternaldocument{supplementary_material}

\usepackage{cvpr}              

\usepackage{graphicx}
\usepackage{amsmath}
\usepackage{amssymb}
\usepackage{booktabs}
\usepackage[linesnumbered,ruled]{algorithm2e}

\usepackage[pagebackref,breaklinks,colorlinks]{hyperref}

\usepackage{enumitem}
\usepackage{caption}
\usepackage{xspace}
\usepackage{amssymb}  
\usepackage{amsmath} 
\usepackage{bm} 
\usepackage{indentfirst} 
\usepackage{tabularx} 
\usepackage{multirow}
\usepackage[table,xcdraw]{xcolor}
\usepackage{booktabs}
\usepackage{hyperref}
\usepackage[linesnumbered,ruled]{algorithm2e}

\usepackage[capitalize]{cleveref}
\crefname{section}{Sec.}{Secs.}
\Crefname{section}{Section}{Sections}
\Crefname{table}{Table}{Tables}
\crefname{table}{Tab.}{Tabs.}

\def\cvprPaperID{8014} 
\def\confName{CVPR}
\def\confYear{2022}

\begin{document}

\title{Revisiting Rolling Shutter Bundle Adjustment:\\Toward Accurate and Fast Solution}
\author{Bangyan Liao$^{1,2,\ast}$ \qquad Delin Qu$^{1,3,\ast}$ \qquad Yifei Xue$^{4}$ \qquad Huiqing Zhang$^{1}$ \qquad Yizhen Lao$^{1,\dagger}$\\
$^{1}$College of Computer Science and Electronic Engineering, Hunan University\\ $^{2}$College of Electrical and Information Engineering, Hunan University \\
$^{3}$Shanghai AI Laboratory\\
$^{4}$Jiangxi Provincial Natural Resources Cause Development Center
}

\maketitle

\begin{abstract}
    We propose an accurate and fast bundle adjustment (BA) solution that estimates the 6-DoF pose with an independent RS model of the camera and the geometry of the environment based on measurements from a rolling shutter (RS) camera. This tackles the challenges in the existing works, namely, relying on high frame rate video as input, restrictive assumptions on camera motion and poor efficiency. To this end, we first verify the positive influence of the image point normalization to RSBA. Then we present a novel visual residual covariance model to standardize the reprojection error during RSBA, which consequently improves the overall accuracy. Besides, we demonstrate the combination of \textbf{N}ormalization and covariance standardization \textbf{W}eighting in \textbf{RSBA} (\textit{NW-RSBA}) can avoid common planar degeneracy without the need to constrain the filming manner. Finally, we propose an acceleration strategy for \textit{NW-RSBA} based on the sparsity of its Jacobian matrix and Schur complement. The extensive synthetic and real data experiments verify the effectiveness and efficiency of the proposed solution over the state-of-the-art works.
    \blfootnote{$\ast$ Authors contributed equally}
    \blfootnote{$\dagger$ Corresponding author: \href{mailto:yizhenlao@hnu.edu.cn}{yizhenlao@hnu.edu.cn}}
    \blfootnote{Project page: \href{https://delinqu.github.io/NW-RSBA}{https://delinqu.github.io/NW-RSBA}}
\end{abstract}

\section{Introduction}
Bundle adjustment (BA) is the problem of simultaneously refining the cameras' relative pose and the observed points' coordinate in the scene by minimizing the reprojection errors over images and points~\cite{MVG}. It has made great success in the past two decades as a vital step for two 3D computer vision applications: structure-from-motion (SfM) and simultaneous localization and mapping (SLAM).

\begin{figure}[t]
    \centering
    \includegraphics[width=.9\linewidth]{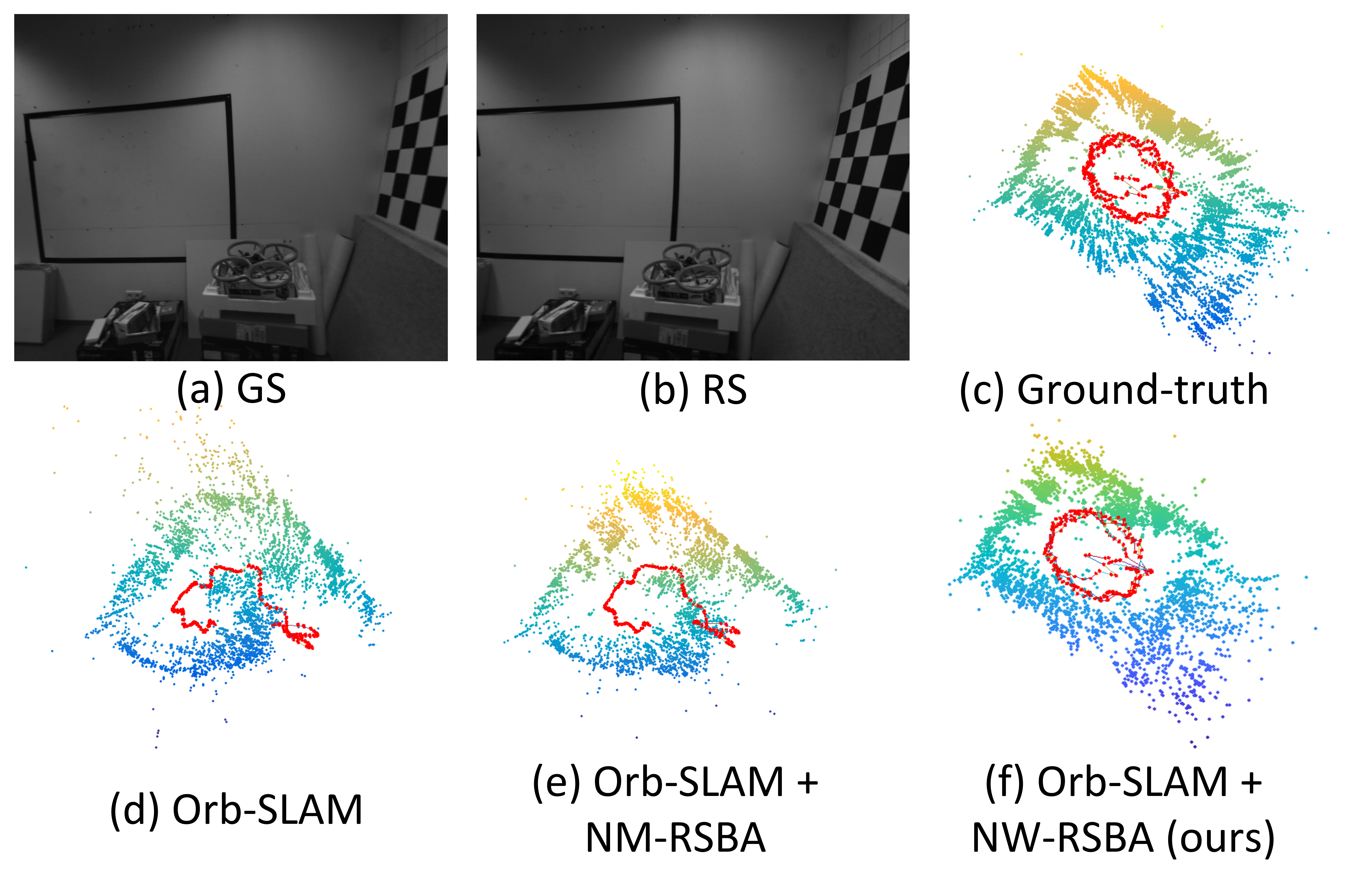}
    \caption{ Images were captured at the same time with fast motion with (a) global shutter and (b) rolling shutter in sequence 10 of the TUM-RSVI dataset~\cite{schubert2019rolling}. (c) classical Orb-SLAM~\cite{mur2015orb} with GS input. (d) classical Orb-SLAM~\cite{mur2015orb} with RS input. (e) Orb-SLAM with \textit{NM-RSBA}~\cite{Albl2016}. (f) Orb-SLAM with proposed \textit{NW-RSBA}.}
    \label{fig:intro}
\end{figure}

The CMOS camera has been widely equipped with the rolling shutter (RS) due to its inexpensive cost, lower energy consumption, and higher frame rate. Compared with the CCD camera and its global shutter (GS) counterpart, RS camera is exposed in a scanline-by-scanline fashion. Consequently, as shown in Fig.~\ref{fig:intro}(a)(b), images captured by moving RS cameras produce distortions known as the RS effect~\cite{Meingast2005}, which defeats vital steps (\textit{e.g.} absolute~\cite{albl2019rolling} and relative~\cite{Dai} pose estimation) in SfM and SLAM, including BA~\cite{Hedborg2012,im2018accurate,Albl2016,ito2016self,lao2018robustified}. Hence, handling the RS effect in BA is crucial for 3D computer vision applications.

\subsection{Related Work}

\noindent \textbf{Video-based Methods.} Hedborg \textit{et al.}~\cite{Hedborg2011} use an RS video sequence to solve RSSfM and present an RSBA algorithm under the smooth motion assumption between consecutive frames in~\cite{Hedborg2012}. Similarly, Im \textit{et al.}~\cite{im2018accurate} propose a small motion interpolation-based RSBA algorithm specifically for narrow-baseline RS image sequences. Zhuang \textit{et al.}~\cite{zhuang2017rolling} further address this setting by presenting 8pt and 9pt linear solvers, which are developed to recover the relative pose of an RS camera that undergoes constant velocity and acceleration motion, respectively. Differently, a spline-based trajectory model to better reformulate the RS camera motion between consecutive frames is proposed by~\cite{patron2015spline}.

\noindent \textbf{Unordered RSBA.} An unordered image set is the standard input for SfM. Albl \textit{et al.}~\cite{Albl2016} address the unordered RSSfM problem and point out the planar degeneracy configuration of RSSfM. Ito \textit{et al.}~\cite{ito2016self} attempt to solve RSSfM by establishing an equivalence with self-calibrated SfM based on the pure rotation instantaneous-motion model and affine camera assumption, while the work of~\cite{lao2021solving} draws the equivalence between RSSfM and non-rigid SfM. A camera-based RSBA  has been proposed in~\cite{lao2018robustified} to simulate the actual camera projection which has the resilience ability against planar degeneracy.

\noindent \textbf{Direct Methods.}  Unlike the feature-based BA methods that minimize reprojection errors of keypoints, photometric-based BA methods maximize the photometric consistency among each pair of frames instead (\textit{e.g.}~\cite{engel2017direct,engel2014lsd}). To handle the RS effect, the works of~\cite{Kim} and~\cite{schubert2018direct} present direct semi-dense and direct spare SLAM pipelines, respectively.

\subsection{Motivations}
\label{section:motivation}

Although existing works of RSBA have shown promising results, we argue that they generally lead to either highly complex constraints on inputs, overly restrictive motion models, or time-consuming which limit application fields.

\noindent \textbf{More general:} \textit{1)} Video-based~\cite{Hedborg2012,im2018accurate,zhuang2017rolling,patron2015spline} and direct methods~\cite{Kim,schubert2018direct} that use video sequence as input are often not desirable to be processed frame by frame which results in high computational power requirements. \textit{2)} The unordered image set is the typical input for classical SfM pipeline (\textit{e.g.}~\cite{wu2011visualsfm}).  \textit{3)} Even the BA step in current popular SLAM systems (\textit{e.g.}~\cite{mur2015orb}) only optimizes keyframes instead of all the consecutive frames.

\noindent \textbf{More effective:} \cite{Albl2016} points out that when the input images are taken with similar readout directions, RSBA fails to recover structure and motion. The proposed solution is simply to avoid the degeneracy configurations by taking images with perpendicular readout directions. This solution considerably limits the use in common scenarios.

\noindent \textbf{More efficient:} GSBA argumentation with the RS motion model has been reported as time-consuming, except for the work of~\cite{Hedborg2012} to accelerate video-based RSBA. However, the acceleration of unordered RSBA has never been addressed in the existing works~\cite{Albl2016,ito2016self,lao2018robustified,lao2021solving}.

\textit{In summary,} an {accurate} and {fast} solution to unordered images RSBA without camera motion assumptions, readout direction is still missing. Such a method would be a vital step in the potential widespread deployment of 3D vision with RS imaging systems.


\subsection{Contribution}
In this paper, we present a novel RSBA solution and tackle the challenges that remained in the previous works. To this end, \textit{1)} we investigate the influence of normalization to the image point on RSBA performance and show its advantages. \textit{2)} Then we present an analytical model for the visual residual covariance, which can standardize the reprojection error during BA, consequently improving the overall accuracy. \textit{3)} Moreover, the combination of \textbf{N}ormalization and covariance standardization \textbf{W}eighting in \textbf{RSBA} (\textit{NW-RSBA}) can avoid common planar degeneracy without constraining the capture manner. \textit{4)} Besides, we propose its acceleration strategy based on the sparsity of the Jacobian matrix and Schur complement. As shown in Fig.~\ref{fig:intro} that \textit{NW-RSBA} can be easily implemented and plugin GSSfM and GSSLAM as  RSSfM and RSSLAM solutions.

In summary, the main contributions of this paper are:

\begin{itemize}[leftmargin=11pt]
    \item We first thoroughly investigate image point normalization's influence on RSBA and propose a probability-based weighting algorithm in the cost function to improve RSBA performance. We apply these two insights in the proposed RSBA framework and demonstrate its acceleration strategy.
    \item The proposed RSBA solution \textit{NW-RSBA} can easily plugin multiple existing GSSfM and GSSLAM solutions to handle the RS effect. The experiments show that \textit{NW-RSBA} provides more accurate results and $10\times$ speedup over the existing works. Besides, it avoids planar degeneracy with the usual  capture manner.
\end{itemize}

\section{Formulation of RSBA}
\label{section：RSBA_formulation}
We formulate the problem of RSBA and provide a brief description of three RSBA methods in existing works. Since this section does not contain our contributions, we give only the necessary details to follow the paper.

\noindent $\bullet$ \textbf{Direct-measurement-based RS model:} Assuming a 3D point $\mathbf{P}_{i} = [X_i \ Y_i \ Z_i]$ is observed by a RS camera $j$ represented by measurement $\mathbf{m}_{i}^j$ in the image domain. The projection from 3D world to the image can be defined as:
\begin{align}
     & 
        {\mathbf{m}_{i}^j}
     = \begin{bmatrix}
        u_i^j & v_i^j
    \end{bmatrix}^\top = \Pi (\mathbf{K}{\mathbf{P}^c}_i^j) \label{RSprojection}, \\
     & {\mathbf{P}^c}_i^j = [{X^c}_i^j \ {Y^c}_i^j \ {Z^c}_i^j]^\top =
    \mathbf{R}^j(v_{i}^j)\mathbf{P}_{i} +  \mathbf{t}^j(v_{i}^j) \label{W2C_transofrm},
\end{align}

\noindent where $\Pi([X \ Y \ Z]^{\top}) = \frac{1}{Z}[X \ Y]^{\top}$ is the perspective division and $\mathbf{K}$ is the camera intrinsic matrix~\cite{MVG}. $\mathbf{R}^j(v_{i}^j) \in \mathbf{SO}({3})$ and $\mathbf{t}^j(v_{i}^j) \in \mathbb{R}^{3}$ define the camera rotation and translation respectively when the row index of measurement $v_{i}^j$ is acquired. Assuming constant camera motion during frame capture, we can model the instantaneous-motion as:
\begin{equation}\label{equation:motion_linear}
    \begin{aligned}
        \resizebox{0.9\linewidth}{!}{
            \begin{math}
                \mathbf{R}^j(v_{i}^j) = (\mathbf{I} + [\boldsymbol{\omega}^j]_{\times} v_{i}^j)\mathbf{R}_{0}^j,\qquad\mathbf{t}^j(v_{i}^j) = \mathbf{t}_{0}^j + \mathbf{d}^j v_{i}^j,
            \end{math}
        }
    \end{aligned}
\end{equation}

\noindent where $[\boldsymbol{\omega}^j]_{\times}$ represents the skew-symmetric matrix of vector $\boldsymbol{\omega}^j$ and $\mathbf{t}_{0}^j$, $\mathbf{R}_{0}^j$ is the translation and rotation matrix when the first row is observed. While $\mathbf{d}^j = [d_{x}^j, d_{y}^j, d_{z}^j]^{\top}$ is the linear velocity vector and $\boldsymbol{\omega}^j = [\omega_{x}^j,\omega_{y}^j,\omega_{z}^j]^{\top}$ is the angular velocity vector. Such model was adopted by~\cite{Dai,Albl2016,lao2021solving}. Notice that RS instantaneous-motion is a function of $v_{i}^j$, which we named the direct-measurement-based RS model.

\noindent $\bullet$ \textbf{Normalized-measurement-based RS model:} By assuming a pre-calibrated camera, one can transform an image measurement $\mathbf{m}_i^j$ with $\mathbf{K}$ to the normalized measurement  $[{\mathbf{q}_i^j}^\top, 1]^\top = \mathbf{K}^{-1} [{\mathbf{m}_i^j}^\top,1]^\top$. Thus, the projection model and camera instantaneous-motion become:
\begin{align}
     & 
        \mathbf{q}_{i}^{j} = \begin{bmatrix}
            c_i^{j} & r_i^{j}
        \end{bmatrix}^\top = \Pi (\mathbf{R}^j(r_{i}^j)\mathbf{P}_{i} +  \mathbf{t}^j(r_{i}^j))\label{equation:projection_nor}, \\
        & \mathbf{R}^j(r_{i}^j) = (\mathbf{I} + [\boldsymbol{\omega}^j]_{\times} r_{i}^j)\mathbf{R}_{0}^j,  \quad\mathbf{t}^j(r_{i}^j) = \mathbf{t}^j_{0} + \mathbf{d}^j r_{i}^j\label{equation:motion_linear_nor}.
\end{align}
In contrast to the direct-measurement-based RS model,  $\mathbf{t}_{0}^j$ and $\mathbf{R}_{0}^j$ are the translation and rotation when the optical centre row is observed. $\boldsymbol{\omega}^j$, $\mathbf{d}^j$ are scaled by camera focal length. We name such model the normalized-measurement-based RS model, which was used in~\cite{Hedborg2012,Albl2016,ito2016self,zhuang2017rolling}.

\noindent $\bullet$ \textbf{Rolling Shutter Bundle Adjustment:} The non-linear least squares optimizers are used to find a solution $\boldsymbol{\theta}^{*}$ including camera poses $\mathbf{R}^{*},\mathbf{t}^{*}$, instantaneous-motion $\boldsymbol{\omega}^{*},\mathbf{d}^{*}$ and 3D points $\mathbf{P}^{*}$  by minimizing the reprojection error $\mathbf{e}_{i}^{j}$ from point $i$ to camera $j$ over all the camera index in set $\mathcal{F}$ and corresponding 3D points index in subset $\mathcal{P}_j$:
\begin{equation}
    \resizebox{0.9\linewidth}{!}{
        \begin{math}
            \begin{aligned}
                 & \boldsymbol{\theta}^* =  \left \{ \mathbf{P}^{*} , \mathbf{R}^{*},\mathbf{t}^{*}, \boldsymbol{\omega}^{*},\mathbf{d}^{*} \right \} = \mathop{\arg\min}_{\boldsymbol{\theta} }  \sum_{j\in \mathcal{F}}\sum_{i\in \mathcal{P}_j }  \left \| \mathbf{e}_{i}^{j} \right \|^{2}_{2}.
            \end{aligned}
        \end{math}
    }
    \label{equation:cost_function}
\end{equation}

\noindent \textit{(1)} \textbf{Direct-measurement-based RSBA:} \cite{duchamp2015rolling} uses the direct-measurement-based
RS model and compute the reprojection error as: $\mathbf{e}_{i}^{j} = \mathbf{m}_{i}^{j} - \Pi (\mathbf{K}(
    \mathbf{R}^j(v_{i}^{j})\mathbf{P}_{i} +  \mathbf{t}^j(v_{i}^{j})
    ))$. We name this strategy \textit{DM-RSBA}.

\noindent \textit{(2)} \textbf{Normalized-measurement-based RSBA:} \cite{Albl2016} uses the normalized-measurement-based RS model and compute the reprojection error as: $\mathbf{e}_{i}^{j} = \mathbf{q}_{i}^{j} -\Pi (
    \mathbf{R}^j(r_{i}^{j})\mathbf{P}_{i} +  \mathbf{t}^j(r_{i}^{j})
    )$. We name this strategy \textit{NM-RSBA}.

\noindent \textit{(3)} \textbf{Direct-camera-base RSBA:} Lao \textit{et al.}~\cite{lao2018robustified} argue both \textit{DM-RSBA} and \textit{NM-RSBA} can not simulate the actual projection. So~\cite{lao2018robustified} proposes a camera-based approach that uses camera pose and instantaneous motion to compute the reprojection without using either $v_i^j$ or $r_i^j$. We name this strategy \textit{DC-RSBA}.


\section{Methodology}

In this section, we present a novel RSBA algorithm called normalized weighted RSBA (\textit{NW-RSBA}). The main idea is to use measurement normalization (section~\ref{sectoin:measurement_normalization}) and covariance standardization weighting (section~\ref{section:NW-RSBA}) jointly during the optimization. Besides, we provide a two-stage Schur complement strategy to accelerate \textit{NW-RSBA} in section~\ref{section:accerlation}. The pipeline of \textit{NW-RSBA} is shown in Alg.~\ref{algorithm:RSBA_pipeline}. The main differences between \textit{NW-RSBA} and existing methods~\cite{duchamp2015rolling,Albl2016,lao2018robustified} are summarized in Tab.~\ref{tab:my-table1}.

\begin{algorithm}[t]
    \SetKwInOut{Input}{Input}
    \SetKwInOut{Output}{Output}

    \Input{Initial rolling shutter camera poses and points as state vector $\boldsymbol{\theta}$. Observed point measurement in normalized image coordinate.}
    \Output{Refined state vector $\boldsymbol{\theta}^{*}$}
    \While{(not reach max iteration) and (not satisfy stopping criteria)}
    {

        \For{ Each camera $j\in \mathcal{F}$}
        {
            \For{ Each point $i\in \mathcal{P}_j$}
            {
                Calculate error $\mathbf{\hat{e}}^j_i $ using Eq.(~\ref{equation:e_hat_definition})\;
                Construct matrix $\mathbf{J}^j_i$ (supplemental material)\;
                Parallel connect $\mathbf{J}^j_i$ to $\mathbf{J}$\;
                Stack $\mathbf{\hat{e}}^{j}_{i}$ into $\mathbf{\hat {e}}$\;
            }

        }
        Solve equation $\mathbf{J}^{\top}\mathbf{J} \boldsymbol{\delta} =-\mathbf{J}^{\top}\mathbf{\hat {e}}$ using Alg.~\ref{algorithm:Series_connection_J}\;
        Update state vector $\boldsymbol{\theta}$ using $\boldsymbol{\delta}$\;
    }
    \caption{Normalized Weighted RSBA}
    \label{algorithm:RSBA_pipeline}
\end{algorithm}

\begin{table}
    \caption{Comparison of the proposed \textit{NW-RSBA} and existing general unordered RSBA methods~\cite{duchamp2015rolling,Albl2016,lao2018robustified}.   D: Direct measurement; N: normalized measurement; M-based: measurement-based projection model; C-based: camera-based projection model.  }
    \centering
    \resizebox{0.48\textwidth}{!}{
        \begin{tabular}{ccccc}
            \toprule
                                                &
            \begin{tabular}[c]{@{}c@{}}\textit{DM-RSBA}\cite{duchamp2015rolling}\end{tabular}          &
            \begin{tabular}[c]{@{}c@{}}\textit{NM-RSBA}\cite{Albl2016}\end{tabular}          &
            \begin{tabular}[c]{@{}c@{}}\textit{DC-RSBA}\cite{lao2018robustified}\end{tabular}          &
            \begin{tabular}[c]{@{}c@{}}\textit{NW-RSBA}\end{tabular}                                                     \\ \hline
            \textbf{Normalization}              & D        & N        & D        & N       \\ \hline
            \textbf{\begin{tabular}[c]{@{}c@{}}Reprojecting \\ computation\end{tabular}} & M-based  & M-based  & C-based  & M-based \\ \hline
            \textbf{\begin{tabular}[c]{@{}c@{}}Analytical \\ Jacobian\end{tabular}} & $\times$ & $\surd$  & $\times$ & $\surd$ \\ \hline
            \textbf{BA acceleration}            & $\times$ & $\times$ & $\times$ & $\surd$ \\ \hline
        \end{tabular}
    }
    \label{tab:my-table1}
\end{table}

\begin{figure}[t]
    \centering
    \includegraphics[width=1\linewidth]{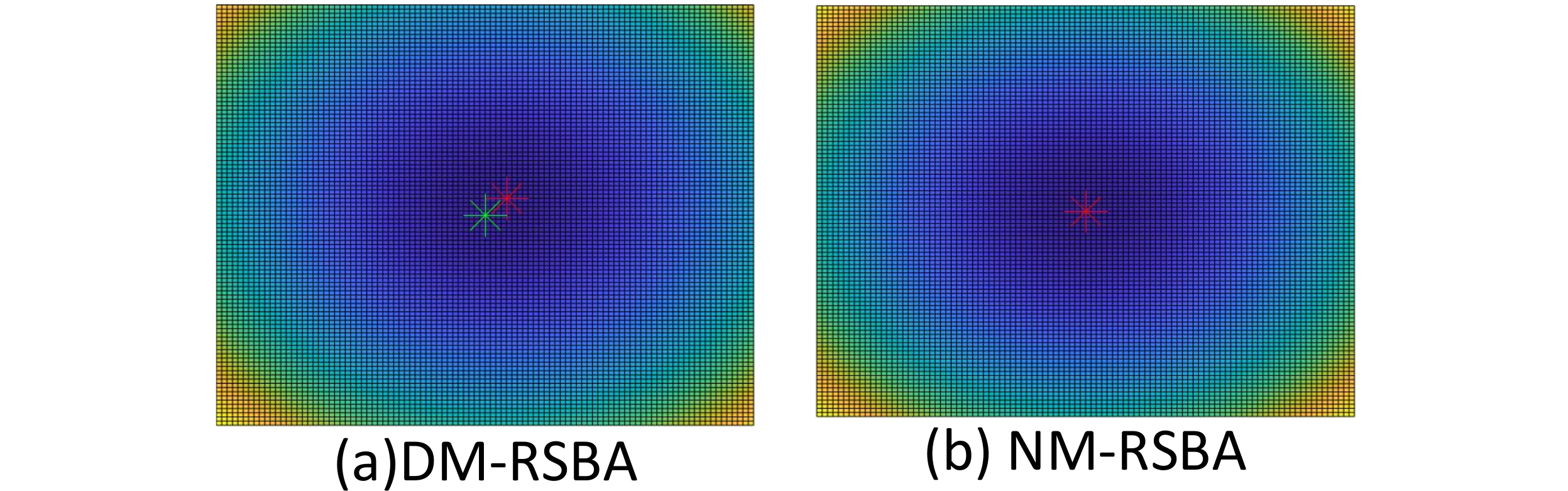}
    \caption{Reprojection loss surface of (a) \textit{DM-RSBA}~\cite{duchamp2015rolling} and (b) \textit{NM-RSBA}~\cite{Albl2016} under the same configuration. The x-axis and y-axis are $d_{x}^j$ and $d_{y}^j$ of camera motion, respectively. The {\color{green}*} are the ground truth solutions while the {\color{red}*} are the minimums in loss surfaces.}
    \label{fig:normalization_loss_surface}
\end{figure}
\subsection{Measurement Normalization}
\label{sectoin:measurement_normalization}

To the best of our knowledge, no prior research has investigated the influence of normalization on  RSBA performance. Thus, we conduct a synthetic experiment to evaluate the impact of normalization by comparing the performances of \textit{DM-RSBA}~\cite{duchamp2015rolling} and \textit{NM-RSBA}~\cite{Albl2016}. The results in Fig.~\ref{fig:synthetic1} show that the normalization significantly improves the RSBA accuracy. The improvement comes from the more precise instantaneous-motion approximation of low-order Taylor expansion in \textit{NM-RSBA}. In \textit{DM-RSBA}, the errors on the image plane induced by the approximate have the same directions and grow exponentially with the increase of the row index. Thus, the optimizer will shift the solution away from the ground truth to equally average the error among all observations. In contrast, the error distribution in \textit{NM-RSBA} is inherently symmetrical due to the opposite direction with the row varying from the optical center, thus exhibiting significantly lower bias. A synthetic example shown in Fig.~\ref{fig:normalization_loss_surface} verifies our insight that \textit{NM-RSBA} has an unbiased local minimum over \textit{DM-RSBA}.

\begin{figure}[t]
    \centering
    \includegraphics[width=1\linewidth]{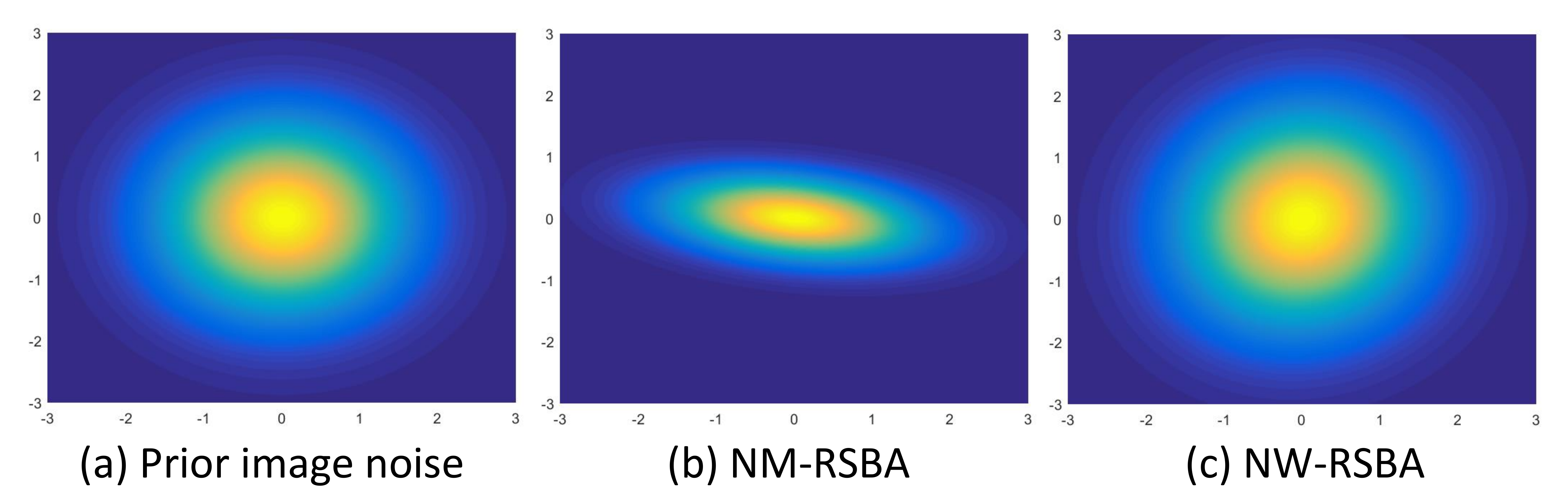}
    \caption{Simulated distributions of (a) prior Gaussian image noise, error term of (b) \textit{NM-RSBA}~\cite{Albl2016}, and the proposed (c) \textit{NW-RSBA}.}
    \label{fig:covariance}
\end{figure}
\subsection{Normalized Weighted RSBA}
\label{section:NW-RSBA}

Based on the measurement normalization, we further present a normalized weighted RSBA (\textit{NW-RSBA}) by modelling image noise covariance.   \\
\noindent $\bullet$ \textbf{Weighting the reprojection error:} In contrast to existing works~\cite{duchamp2015rolling,Albl2016,lao2018robustified}, we consider the image noise in RSBA by weighting the squared reprojection error with the inverse covariance matrix of the error, which is also known as standardization of the error terms~\cite{triggs1999bundle}. Thus the cost function in Eq.~(\ref{equation:cost_function}) becomes:
\begin{equation}
    \begin{aligned}
        \boldsymbol{\theta}^*  = \mathop{\arg\min}_{\boldsymbol{\theta} }  \sum_{j\in \mathcal{F}}\sum_{i\in \mathcal{P}_j }  \textrm{}{\mathbf{e}_{i}^{j}}^\top {\mathbf{\Sigma}_{i}^{j}}^{-1} {\mathbf{e}_{i}^{j}},
    \end{aligned}
    \label{equation:cost_function_NW}
\end{equation}
\noindent where $\mathbf{e}_{i}^{j}$ is the reprojection error computed by normalized-measurement-based approach described in Eq.~(\ref{equation:projection_nor}). By assuming the image measurement $\mathbf{q}_i^j$ follows a prior Gaussian noise: $\mathbf{n}_i^j \sim \mathcal{N}(0, \mathbf{W\Sigma W}^{\top})$, the newly introduced covariance matrix of the error ${\mathbf{\Sigma}_{i}^{j}}$ is defined as follows (proof in the supplemental material):
\begin{equation}
\begin{aligned}
&\mathbf{\Sigma}_{i}^{j} = \mathbf{C}_i^j \mathbf{W} \mathbf{\Sigma}\mathbf{W}^{\top} {\mathbf{C}_i^j}^\top, \label{equation:sigma}
\end{aligned}
\end{equation}
\begin{equation}\label{equation:sigma_c}
    \resizebox{0.90\linewidth}{!}{
        \begin{math}
            \begin{aligned}
                \mathbf{C}_i^j = \begin{bmatrix}1 &0 \\ 0 &1\end{bmatrix} - \begin{bmatrix}
                    \frac{1}{{{Z^c}_i^j}}              & 0                                  \\
                    0                                  & \frac{1}{{{Z^c}_i^j}}              \\
                    \frac{-{{X^c}_i^j}}{{{Z^c}_i^j}^2} & \frac{-{{Y^c}_i^j}}{{{Z^c}_i^j}^2}
                \end{bmatrix}^\top ([\boldsymbol{\omega}^j]_\times\mathbf{R}^j\mathbf{P}_i+\mathbf{d}^j) {\begin{bmatrix}0 \\1 \end{bmatrix}}^\top,
            \end{aligned}
        \end{math}
    }
\end{equation}

\begin{equation}\label{equation:sigma_w}
    \begin{aligned}
        \mathbf{W} =  \begin{bmatrix}
            1/f_x & 0     \\
            0     & 1/f_y
        \end{bmatrix},
    \end{aligned}
\end{equation}

\noindent where $\mathbf{C}_i^j$ and $\mathbf{W}$ are  $2\times2$ auxiliary matrices. $f_x$ and $f_y$ are focal lengths.

\begin{figure}[t]
    \centering
    \includegraphics[width=1\linewidth]{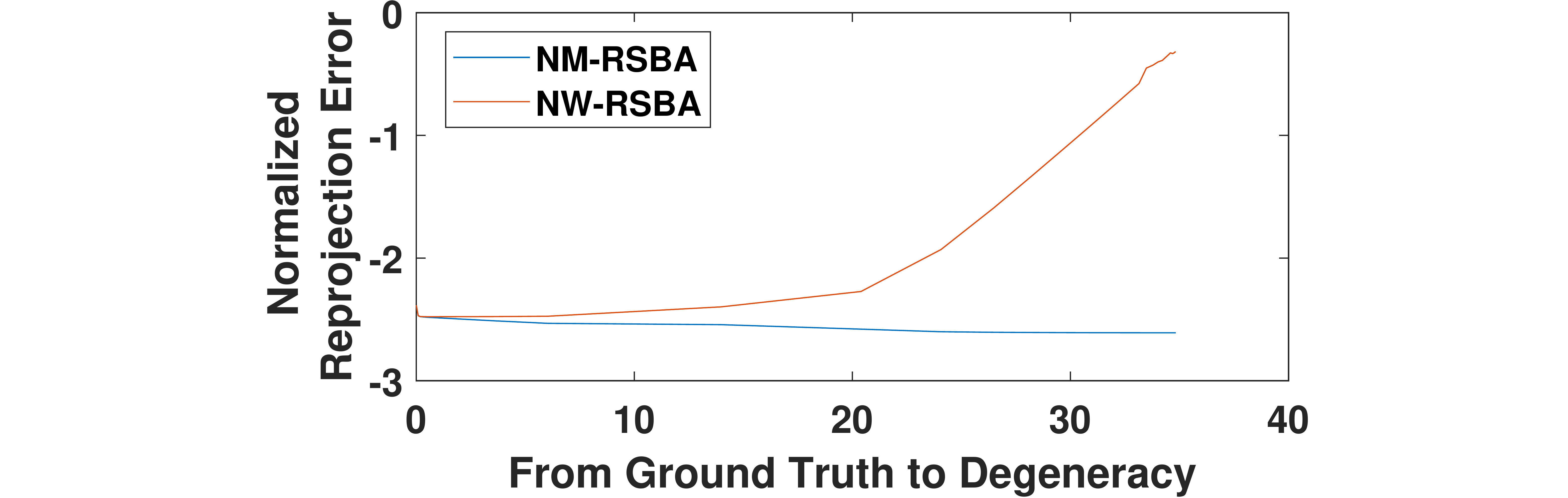}

    \caption{The normalized reprojection error comparison between our proposed method \textit{NW-RSBA} and \textit{NM-RSBA}~\cite{Albl2016} along the degeneracy process.}
    \label{fig:loss_surface}
\end{figure}

\begin{figure}[t]
    \centering
    \includegraphics[width=1\linewidth]{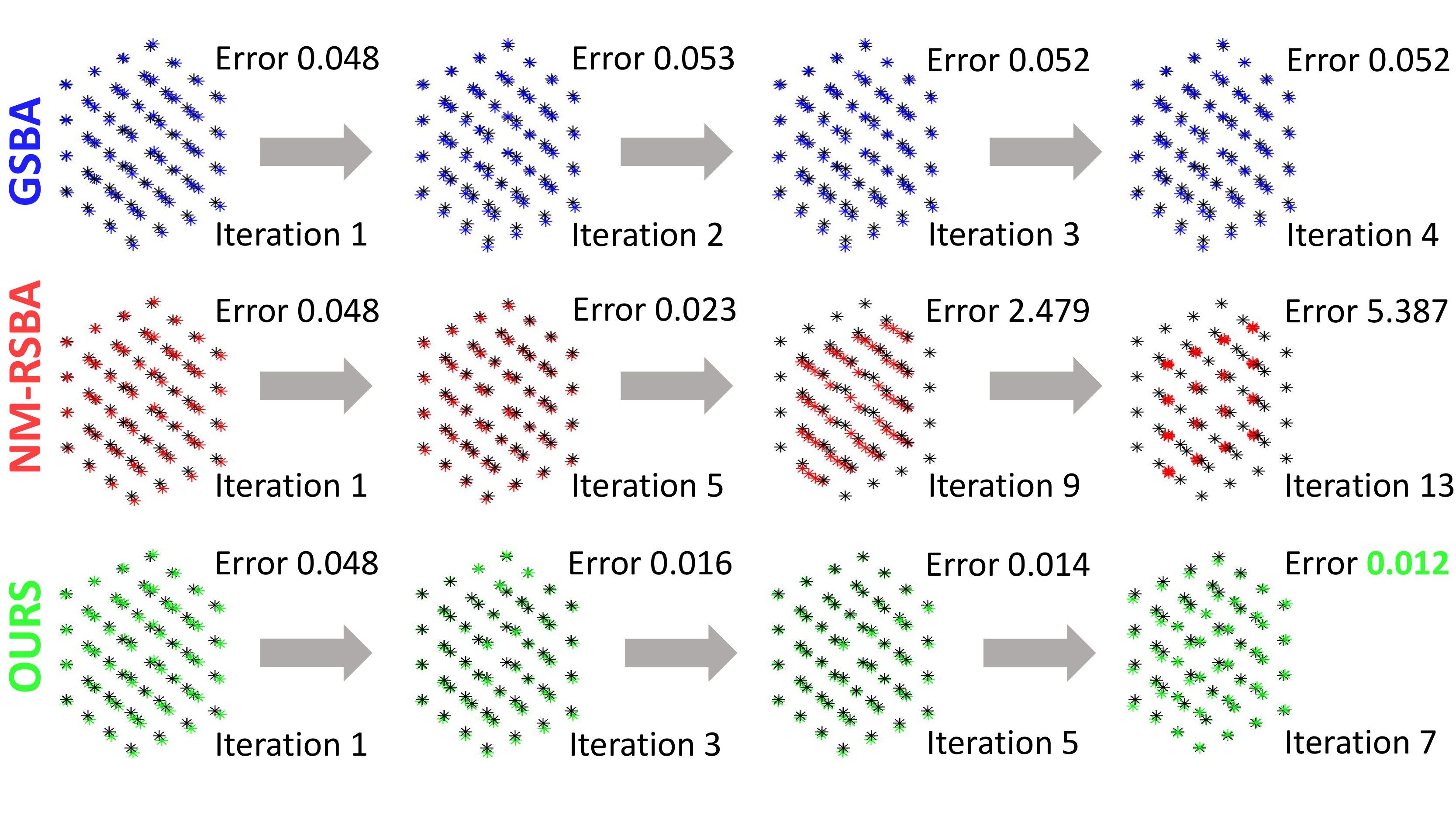}
    \caption{ Analysis of the degeneracy in synthetic cube scene captured by five cameras with parallel readout direction. The reconstruction of \textit{NM-RSBA}~\cite{Albl2016} (\textcolor{red}{red}) collapses to the plane gradually while \textit{NW-RSBA} (\textcolor{green}{green}) provides the most accurate result.}
    \label{fig:degeneracy_test}
\end{figure}

\noindent $\bullet$ \textbf{Advantages of noise covariance weighting:}  Note that the  standardisation in Eq.~(\ref{equation:cost_function_NW}) scaling every term by its inverse covariance matrix ${\mathbf{\Sigma}_{i}^{j}}^{-1}$, so that all terms end up with isotropic covariance~\cite{triggs1999bundle}. In Fig.~\ref{fig:covariance}, we visualize the influence of error terms standardization, which re-scales the prior covariance ellipse back to a unit one. We interpret the re-weighting standardization leads two advantages:

\noindent \textit{(1)} \textbf{More accurate:} With the re-weighting standardization in Eq.~(\ref{equation:cost_function_NW}),   features with a high variance which means a high probability of a large reprojection error, are offered less confidence to down-weighting their influence on the total error. Synthetic experiments in section~{\ref{section：Synthetic_Data}} demonstrate it outperforms \textit{NM-RSBA} under various noise levels.

\noindent \textit{(2)}  \textbf{Handle planar degeneracy:} Though the proposed \textit{NW-RSBA} uses measurement-based projection during degeneracy, it still provides stable BA results and even outperforms \textit{C-RSBA} with the noise covariance weighting. As demonstrated in~\cite{Albl2016}, under the planar degeneracy configuration, the y-component of the reprojection error will reduce to zero, which denotes that the covariance matrix holds a zero element in the y direction. \textit{NW-RSBA} explicitly modeled the error covariance and standardized it to isotropy. Thus, the proposed method exponentially amplifies the error during degeneracy, as shown in Fig.~\ref{fig:loss_surface}, and prevents the continuation of the decline from ground truth to the degenerated solution (proofs can be found in the supplemental material). A synthetic experiment shown in Fig.~\ref{fig:degeneracy_test} verifies that \textit{NM-RSBA} easily collapses into degeneracy solutions while \textit{NW-RSBA} provide stable result and outperforms the \textit{GSBA}. 

\noindent $\bullet$ \textbf{Jacobian of \textit{NW-RSBA}:} For more convenient optimization, we reformulate covariance matrix $\mathbf{\Sigma}_{i}^{j}$ as a standard least square problem using decomposition:
\begin{equation}
    {\mathbf{\Sigma}_{i}^{j}}^{-1} = {\mathbf{C}_i^j}^{-\top}\mathbf{W}^{-\top}\mathbf{\Sigma}^{-\frac{1}{2}}\mathbf{\Sigma}^{-\frac{1}{2}}\mathbf{W}^{-1}{\mathbf{C}_i^j}^{-1}.
    \label{equation:covar_decomposition}
\end{equation}
By substituting Eq.~(\ref{equation:covar_decomposition}) into~(\ref{equation:cost_function_NW}), we get a new cost function:
\begin{equation}
    \boldsymbol{\theta}^* = \mathop{\arg\min}_{\boldsymbol{\theta}}  \sum_{j\in \mathcal{F}}\sum_{i\in \mathcal{P}_j } \textrm{}{{\mathbf{\hat{e}}^{j\top}_i}} {\mathbf{\hat{e}}^j_i},
    \label{equation:Jac1}
\end{equation}
\begin{equation}\label{equation:e_hat_definition}
    \noindent \text{where},
    \begin{aligned}
        {\mathbf{\hat{e}}_i^j} = \mathbf{\Sigma}^{-\frac{1}{2}}\mathbf{W}^{-1}{\mathbf{C}_i^j}^{-1}\mathbf{e}_i^j.
    \end{aligned}
\end{equation}
We derive the analytical Jacobian matrix of ${\mathbf{\hat{e}}^j_i}$ in Eq.~(\ref{equation:Jac1}) using the chain rule in the supplemental material.

\begin{figure}[t]
    \begin{center}
        \includegraphics[width=.9\linewidth]{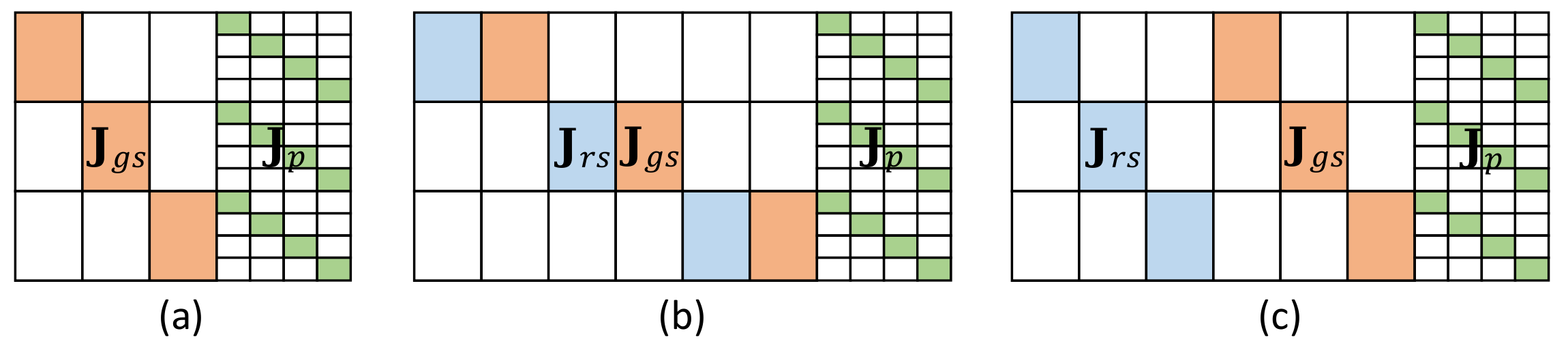}
    \end{center}
    \caption{Example Jacobian matrices with 4 points
        and 3 cameras in (a) GSBA, RSBA with (b) series and (c) parallel connection. }
    \label{fig:BA_J}
\end{figure}

\begin{figure}[t]
    \centering
    \includegraphics[width=1\linewidth]{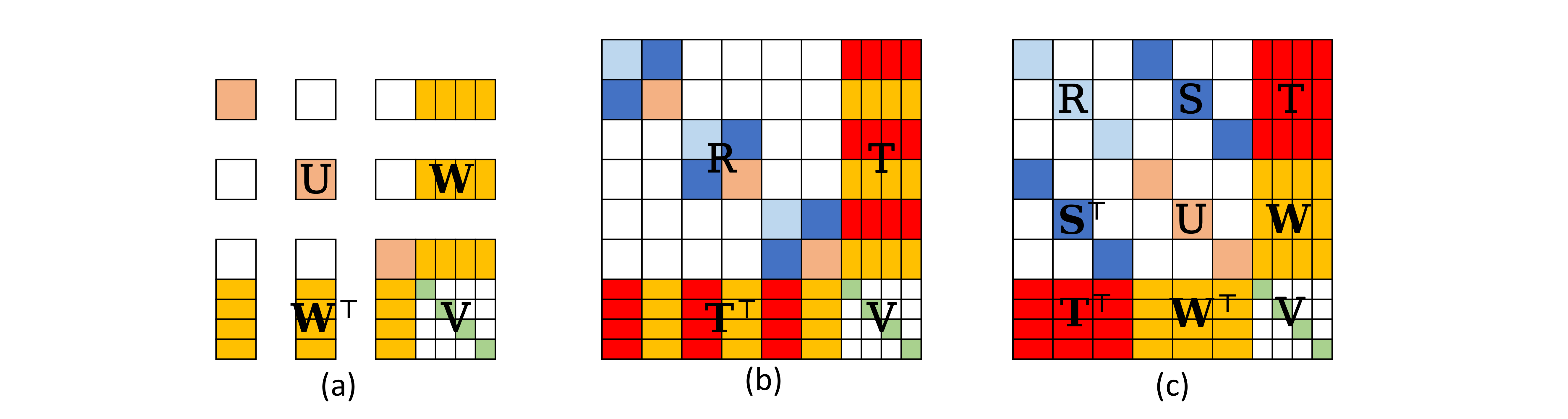}
    \caption{Example Hessian matrices with 4 points
        and 3 cameras in (a) GSBA, RSBA with(b) series and (c) parallel connection. }
    \label{fig:Hessien_matrix}
\end{figure}

\begin{algorithm}[t]
    \SetKwInOut{Input}{Input}
    \SetKwInOut{Output}{Output}

    \Input{Jacobian matrix $\mathbf{J}$ and weighted error vector $\mathbf{\hat{e}}$}
    \Output{Update state vector $\boldsymbol{\delta}$}
    Compute Schur complement matrix $\textbf{S}_p$ and $\textbf{S}_{rs}$  using Eq.~(\ref{equation:Sp}) and (\ref{equation:S_gs})\;
    Compute auxiliary vectors $\textbf{t}^{*}$ and $\textbf{u}^{*}$using Eq.~(\ref{equation:t_new}) and (\ref{equation:u_new})\;
    Solve Eq.~(\ref{equation:new_normal_equation}) cascadingly:
    \begin{itemize}
        \item Get $\boldsymbol{\delta}_{rs}$ by solving $\mathbf{S}_{rs}\boldsymbol{\delta}_{rs} = -\mathbf{t}^{*}$\;
        \item Get $\boldsymbol{\delta}_{gs}$ by solving $\mathbf{U}^{*} \boldsymbol{\delta}_{gs}= - \textbf{u}^{*} - \mathbf{S}^{*\top}\boldsymbol{\delta}_{rs}$\;
        \item Get $\boldsymbol{\delta}_{p}$ by solving $\mathbf{V} \boldsymbol{\delta}_{p}= - \mathbf{v} - \mathbf{T}^{\top}\boldsymbol{\delta}_{rs} - \mathbf{W}^{\top} \boldsymbol{\delta}_{gs}$\;

        \item Stack $\boldsymbol{\delta}_{gs}$,  $\boldsymbol{\delta}_{rs}$,  $\boldsymbol{\delta}_{p}$ into $\boldsymbol{\delta}$\;
    \end{itemize}

    \caption{Solve the normal equation using two-stage Schur complement}
    \label{algorithm:Series_connection_J}
\end{algorithm}

\subsection{NW-RSBA Acceleration}
\label{section:accerlation}
Based on the sparsity of the Jacobian, the marginalization~\cite{triggs1999bundle,lourakis2009sba} with Schur complement has achieved significant success in accelerating GSBA. However, the acceleration strategy has never been addressed for the general unordered RSBA  in~\cite{duchamp2015rolling,Albl2016,lao2018robustified}. As shown in Fig.~\ref{fig:BA_J}(b)(c), we can organize the RSBA Jacobian in two styles:

\noindent \textit{(1)} \textbf{Series connection:} By connecting camera pose and instantaneous-motion in the Jacobian matrix (Fig.~\ref{fig:Hessien_matrix}(b)) as an entirety, we can use the \textbf{one-stage Schur complement} technique~\cite{lourakis2009sba} to marginalize out the 3D point and compute the update state vector for $\mathbf{R},\mathbf{t},\boldsymbol{\omega},\mathbf{d}$ first, followed by back substitution for update state vector of points $\mathbf{P}$.

\noindent \textit{(2)} \textbf{Parallel connection:} Due to the independence between camera pose and instantaneous-motion in the Jacobian matrix (Fig.~\ref{fig:Hessien_matrix}(c)), we propose a \textbf{two-stage Schur complement} strategy to accelerate RSBA. When solving the non-linear least square problem (\textit{e.g.} Gauss-Newton), the approximate Hessian matrix for Eq.~(\ref{equation:Jac1}) is defined as
\begin{equation}
    \resizebox{0.85\linewidth}{!}{
        \begin{math}
            \begin{aligned}
                \textbf{J}^{\top}\textbf{J} =
                \begin{bmatrix}{
                    \textbf{J}_{rs}}^\top{\textbf{J}_{rs}}  & {\textbf{J}_{rs}}^\top \textbf{J}_{gs} & {\textbf{J}_{rs}}^\top \textbf{J}_{p} \\
                    {\textbf{J}_{gs}}^\top{\textbf{J}_{rs}} & {\textbf{J}_{gs}}^\top \textbf{J}_{gs} & {\textbf{J}_{gs}}^\top \textbf{J}_{p} \\
                    {\textbf{J}_{p}}^\top{\textbf{J}_{rs}}  & {\textbf{J}_{p}}^\top \textbf{J}_{gs}  & {\textbf{J}_{p}}^\top \textbf{J}_{p}  \\
                \end{bmatrix} =
                \begin{bmatrix}
                    \textbf{R}        & \textbf{S}        & \textbf{T} \\
                    \textbf{S}^\top   & \textbf{U}        & \textbf{W} \\
                    {\textbf{T}^\top} & {\textbf{W}^\top} & \textbf{V}
                \end{bmatrix},
            \end{aligned}
        \end{math}
    }
    \label{equation:Hessian_matrix}
\end{equation}
\begin{align}
    \mathbf{J}       & = \begin{bmatrix} \mathbf{J}_{rs} & \mathbf{J}_{gs} & \mathbf{J}_p \end{bmatrix}
    = \begin{bmatrix}
        \frac{\partial {\mathbf{\hat{e}}}}{\partial \mathbf{x}_{rs}} & \frac{\partial {\mathbf{\hat{e}}}}{\partial \mathbf{x}_{gs}} & \frac{\partial {\mathbf{\hat{e}}}}{\partial \mathbf{x}_{p}}
    \end{bmatrix}, \tag{\ref{equation:Hessian_matrix}{a}} \label{Hessian_matrix_supp1} \\
    \mathbf{\hat{e}} & = \left [ \begin{Bmatrix}
            \mathbf{\hat{e}}_i^j
        \end{Bmatrix} \right ], \quad
    \mathbf{x}_{gs} = \begin{bmatrix}
        \begin{Bmatrix}
            \mathbf{R}^j
        \end{Bmatrix} & \begin{Bmatrix}
            \mathbf{t}^j
        \end{Bmatrix}
    \end{bmatrix}
    , \tag{\ref{equation:Hessian_matrix}{b-1}} \label{Hessian_matrix_supp2}                           \\
    \mathbf{x}_{rs}  & = \begin{bmatrix}
        \begin{Bmatrix}
            \boldsymbol{\omega}^j
        \end{Bmatrix} & \begin{Bmatrix}
            \mathbf{d}^j
        \end{Bmatrix}
    \end{bmatrix}, \quad
    \mathbf{x}_{p} = \left [   \begin{Bmatrix}
            \mathbf{P}_i
        \end{Bmatrix} \right ], \tag{\ref{equation:Hessian_matrix}{b-2}} \label{Hessian_matrix_supp3}
\end{align}

\noindent where $\mathbf{R}$, $\mathbf{U}$, $\mathbf{V}$, $\mathbf{S}$, $\mathbf{T}$ and $\mathbf{W}$ are submatrices computed by the derivations $\mathbf{J}_{rs}$, $ \mathbf{J}_{gs}$ and $ \mathbf{J}_{p}$. As Alg.~\ref{algorithm:Series_connection_J} shown that the two-stage Schur complement strategy consists of 3 steps:

\noindent \textbf{$\triangleright $ Step 1: Construct normal equation.} In each iteration,  using this form of the Jacobian and corresponding state vectors and the error vector, the normal equation follows

\begin{equation}
    \resizebox{0.85\linewidth}{!}{
        \begin{math}
            \begin{aligned}
                \mathbf{J}^{\top}\mathbf{J} {\boldsymbol{\delta}}=
                \begin{bmatrix} \mathbf{R} & \mathbf{S} & \mathbf{T}\\ \mathbf{S}^\top &\mathbf{U} &\mathbf{W}\\ \mathbf{T}^\top & \mathbf{W}^\top & \mathbf{V} \end{bmatrix}\begin{bmatrix}
                    \boldsymbol{\delta}_{rs} \\
                    \boldsymbol{\delta}_{gs} \\
                    \boldsymbol{\delta}_{p}
                \end{bmatrix}
                 = -\begin{bmatrix}
                    \mathbf{t} \\
                    \mathbf{u} \\
                    \mathbf{v}
                \end{bmatrix}=-\mathbf{J}^\top \mathbf{\hat{e}},
            \end{aligned}
        \end{math}
    }
    \label{equation:normal_equation}
\end{equation}

\noindent where $\boldsymbol{\delta}_{gs}$, $\boldsymbol{\delta}_{rs}$ and $\boldsymbol{\delta}_{p}$ are the update state vectors to $\mathbf{x}_{gs}$, $\mathbf{x}_{rs}$, and $\mathbf{x}_{p}$, while $\mathbf{t}$, $\mathbf{u}$, and $\mathbf{v}$ are the corresponding descent direction. Such formed normal equations show a block sparsity, suggesting that we can efficiently solve it.

\noindent \textbf{$\triangleright $ Step 2: Construct Schur complement.}  We construct two-stage Schur complements $\mathbf{S}_p$, $\mathbf{S}_{rs}$ and two auxiliary vectors $\mathbf{u}^{*}$ and $\mathbf{t}^{*}$ to Eq.~(\ref{equation:normal_equation}) as
\begin{equation}
    \resizebox{0.85\linewidth}{!}{
    \begin{math}
    \begin{aligned}
     \mathbf{S}_p = \begin{bmatrix}\mathbf{R}^* & \mathbf{S}^{*}\\ \mathbf{S}^{*\top} & \mathbf{U}^*\end{bmatrix}
        = \begin{bmatrix} \mathbf{R} - \mathbf{T}\mathbf{V}^{-1}\mathbf{T}^\top &  \mathbf{S} - \mathbf{T}\mathbf{V}^{-1}\mathbf{W}^\top \\ \mathbf{S}^\top - \mathbf{W}\mathbf{V}^{-1}\mathbf{T}^\top &  \mathbf{U} - \mathbf{W}\mathbf{V}^{-1}\mathbf{W}^\top\end{bmatrix},  \\ \label{equation:Sp}                                                                                         
    \end{aligned}
    \end{math}
    }
\end{equation}
\vspace{-5.5ex}
\begin{align} 
 \mathbf{S}_{rs} &= \mathbf{R}^* - \mathbf{S}^{*} {\mathbf{U}^{*}}^{-1}{\mathbf{S}^{*}}^{\top}, \label{equation:S_gs}                     \\
 \mathbf{u}^{*} &= \mathbf{u} - \mathbf{W}\mathbf{V}^{-1}\mathbf{v}\label{equation:u_new}, \\
 \textbf{t}^{*} &= \textbf{t} - \textbf{T}\textbf{V}^{-1}\textbf{v} - \textbf{S}^{*} \textbf{U}^{*-1}\textbf{u}^* \label{equation:t_new}.
\end{align}

\noindent \textbf{$\triangleright $ Step 3: Orderly solve $\boldsymbol{\delta}_{gs}$, $\boldsymbol{\delta}_{rs}$ and $\boldsymbol{\delta}_{p}$.} Based on  $\mathbf{S}_p$, $\mathbf{S}_{rs}$, $\mathbf{u}^{*}$ and $\mathbf{t}^{*}$, we reformulate the normal equation as

\begin{equation}
    \begin{aligned}
        \begin{bmatrix}\mathbf{S}_{rs}    & \mathbf{0}        & \mathbf{0}   \\
            \mathbf{S}^{*\top} & {\mathbf{U}^*}    & \mathbf{0}   \\
            {\mathbf{T}^\top}  & {\mathbf{W}^\top} & {\mathbf{V}}\end{bmatrix}
        \begin{bmatrix}
            \boldsymbol{\delta}_{rs} \\
            \boldsymbol{\delta}_{gs} \\
            \boldsymbol{\delta}_{p}
        \end{bmatrix}
        = -  \begin{bmatrix}
            \mathbf{t}^{*} \\
            \mathbf{u}^*   \\
            \mathbf{v}
        \end{bmatrix},
    \end{aligned}
    \label{equation:new_normal_equation}
\end{equation}

\noindent which enables us to compute $\boldsymbol{\delta}_{rs}$ first, and then back substitutes the results to get $\boldsymbol{\delta}_{gs}$. Finally, we can obtain $\boldsymbol{\delta}_{p}$ based on the $3^{\text{rd}}$ row of Eq.~(\ref{equation:new_normal_equation}).

\subsection{Implementation}
We follow Alg.~\ref{algorithm:RSBA_pipeline} to implement the proposed \textit{NW-RSBA} in C++. The implemented \textit{NW-RSBA} can serve as a little module and can be easily plug-in such context:

\noindent \textbf{$\bullet$ RS-SfM:}  We augment VisualSFM~\cite{wu2011visualsfm} by shifting the incremental GSBA pipeline with the proposed \textit{NW-RSBA}.

\noindent \textbf{$\bullet$ RS-SLAM:} We augment Orb-SLAM~\cite{mur2015orb} by replacing the local BA and full BA modules with  \textit{NW-RSBA}.

\begin{figure}
    \centering
    \includegraphics[width=1\linewidth]{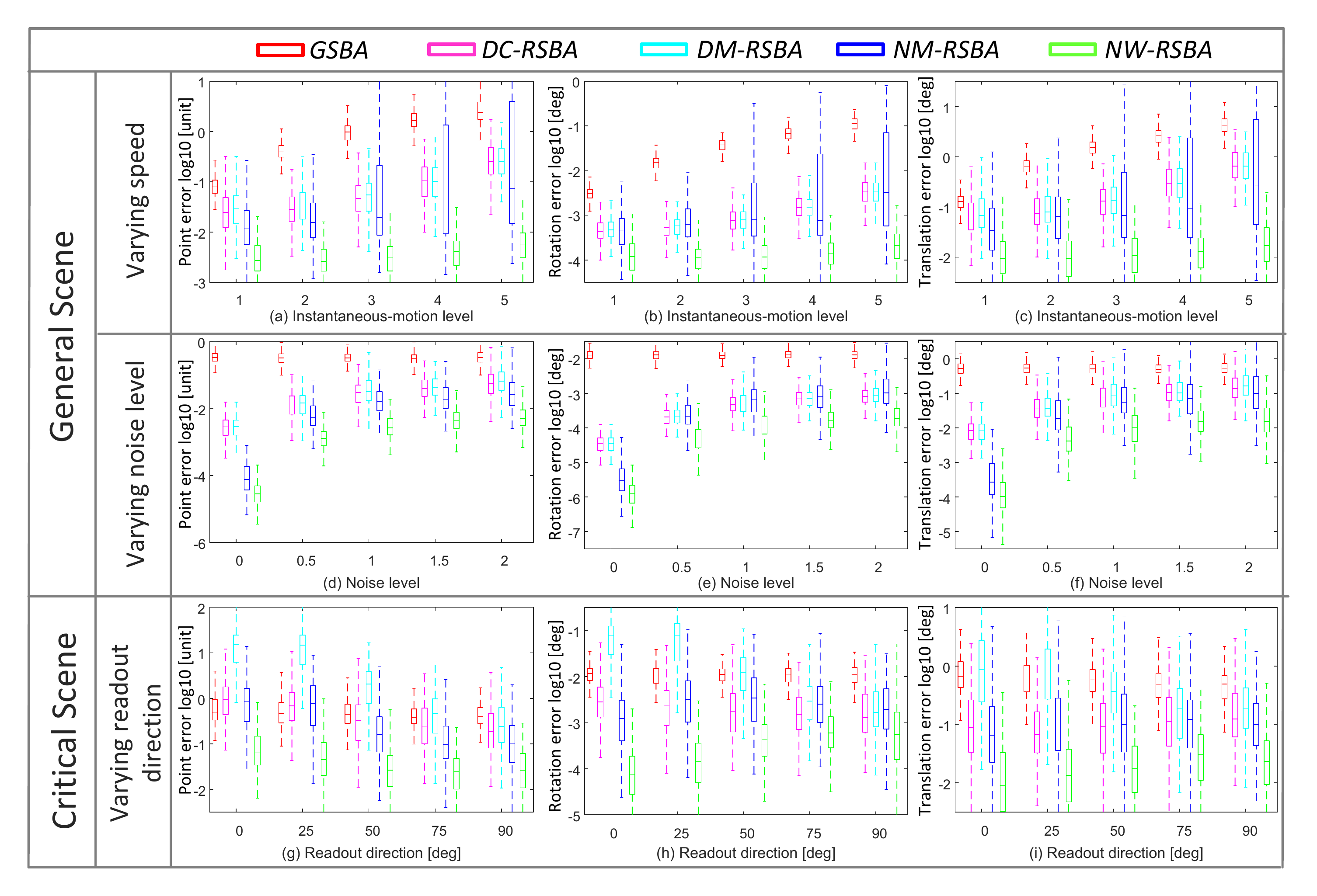}
    \caption{Camera pose ($2^\text{nd}$ and $3^\text{rd}$ columns) and reconstruction ($1^\text{st}$ column) errors of \textit{GSBA}, \textit{DC-RSBA}, \textit{DM-RSBA}, \textit{NM-RSBA} and \textit{NW-RSBA} with increasing angular and linear velocity  ($1^\text{st}$ row) and noise levels in the image ($2^\text{nd}$ row) in a general scene, also with increasing readout directions in a degeneracy scene ($3^\text{rd}$ row).}
    \label{fig:synthetic1}
    \vspace{-2ex}
\end{figure}
\begin{figure}
    \centering
    \includegraphics[width=1\linewidth]{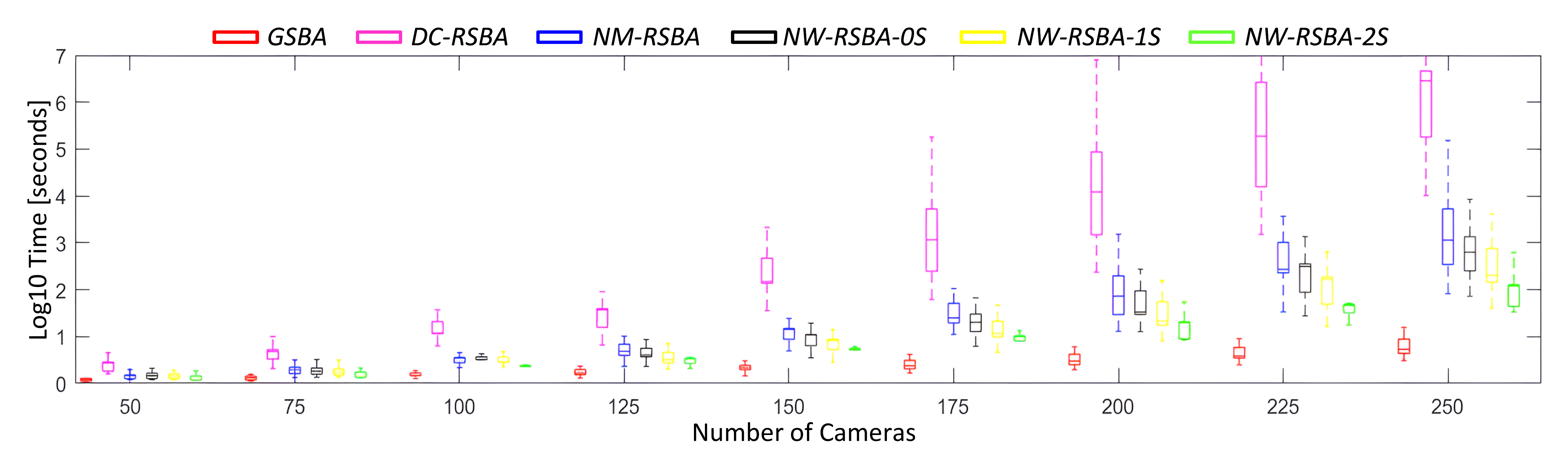}
    \caption{Time cost of \textit{GSBA}~\cite{lourakis2009sba}, \textit{DC-RSBA}~\cite{lao2018robustified}, \textit{NM-RSBA}~\cite{Albl2016}, \textit{NW-RSBA-0S} (without Schur complement), \textit{NW-RSBA-1S} (one-stage Schur complement to Jacobian matrices with series connection), and proposed \textit{NW-RSBA-2S} (two-stage Schur complement to Jacobian matrices with parallel connection) with increasing camera number and fixed point number.}
    \label{fig:time}
\end{figure}
\begin{figure*}
    \centering
    \includegraphics[width=.8\linewidth]{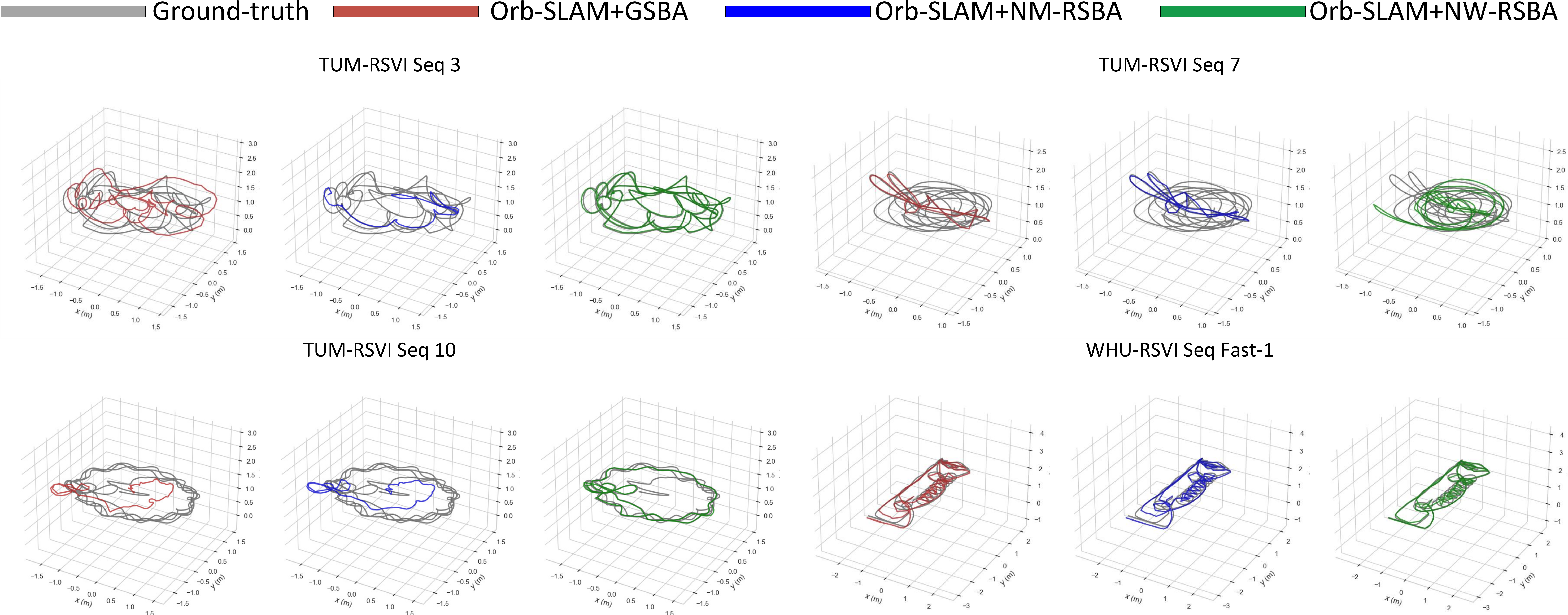}
    \caption{Ground truth and trajectories estimated by \textit{GSBA}~\cite{lourakis2009sba}, \textit{NM-RSBA}~\cite{Albl2016} and proposed \textit{NW-RSBA} after Sim(3) alignment on 3 sequences from TUM-RSVI~\cite{schubert2019rolling} and 1 sequence from WHU-RSVI~\cite{cao2020whu} datasets.}
    \label{fig:real_slam_traj}
\end{figure*}

\section{Experimental Evaluation}
In our experiments, the proposed method is compared to three state-of-the-art unordered RSBA solutions:\textit{1) GSBA}: SBA~\cite{lourakis2009sba}.
\textit{2) DC-RSBA}: direct camera-based RSBA~\cite{lao2018robustified}.
\textit{3) DM-RSBA}: direct measurement-based RSBA~\cite{duchamp2015rolling}.
\textit{4) NM-RSBA}: normalized-measurement-based RSBA~\cite{Albl2016}.
\textit{5) NW-RSBA}: proposed normalized weighted RSBA.

\subsection{Synthetic Data}
\label{section：Synthetic_Data}
\noindent \textbf{Settings and metrics.} We simulate $5$ RS cameras located randomly on a sphere pointing at a cubical scene. We compare all methods by varying the speed, the image noise, and the readout direction. The results are obtained after collecting the errors over 300 trials per epoch. We measure the reconstruction errors and pose errors.

\noindent \textbf{Results.}
\textit{1) \textbf{Varying Speed.}} The results in Fig.~\ref{fig:synthetic1}(a)(b)(c) show that the estimated errors of \textit{GSBA} grow with speed  while \textit{DC-RSBA}, \textit{DM-RSBA} and \textit{NM-RSBA}  achieve better results with slow kinematics. The proposed \textit{NW-RSBA} provides the best results under all configurations.
\textit{2) \textbf{Varying Noise Level.}}  In Fig.~\ref{fig:synthetic1}(d)(e)(f), \textit{GSBA} shows better robustness to noise but with lower accuracy than RS methods. The proposed \textit{NW-RSBA} achieves the best performance with all noise levels.
\textit{3) \textbf{Varying Readout Direction.}} We evaluate five methods with varying readout directions of the cameras by increasing the angle from parallel to perpendicular.  Fig.~\ref{fig:synthetic1}(g)(h)(i) show that  under a small angle, the reconstruction error of \textit{DM-RSBA}, \textit{DC-RSBA} and \textit{DM-RSBA}  grow dramatically even bigger than \textit{GSBA}, suggesting a degenerated solution.
In contrast, \textit{NW-RSBA} provides stable results under all settings, even with the parallel readout direction.

\noindent \textbf{Runtime.} As shown in Fig.~\ref{fig:time} that without analytical Jacobian, \textit{DC-RSBA} is the slowest one while the proposed \textit{NW-RSBA} achieves similar efficiency as \textit{NM-RSBA}. However, by using acceleration strategies, \textit{NW-RSBA-1S} and \textit{NW-RSBA-2S} reduce the overall runtime. Note that \textit{NW-RSBA-2S} achieves an order of magnitude faster than \textit{NM-RSBA}.

\subsection{Real Data}
\noindent \textbf{Datasets and metrics.} We compare all the RSBA methods in two publicly available RS datasets: WHU-RSVI~\cite{cao2020whu} dataset\footnote{\href{http://aric.whu.edu.cn/caolike/2019/11/05/the-whu-rsvi-dataset}{http://aric.whu.edu.cn/caolike/2019/11/05/the-whu-rsvi-dataset}}, TUM-RSVI~\cite{schubert2019rolling} dataset\footnote{\href{https://vision.in.tum.de/data/datasets/rolling-shutter-dataset}{https://vision.in.tum.de/data/datasets/rolling-shutter-dataset}}. In this section, we use three evaluation metrics, namely ATE $e_{\text{ate}}$ (absolute trajectory error)~\cite{schubert2019rolling}, tracking duration $\underline{DUR}$ (the ratio of the successfully tracked frames out of the total frames) and real-time factor $\epsilon$ (sequence's actual duration divided by the algorithm's processing time).

\begin{table}[]
    \caption[]{ Statistics of 10 sequences from TUM-RSVI~\cite{schubert2019rolling} and 2 sequences from WHU-RSVI~\cite{cao2020whu} datasets. The realtime factor $\epsilon$ and tracking duration $\underline{DUR}$ of  \textit{Orb-SLAM}, \textit{Orb-SALM+NM-RSBA}~\cite{Albl2016}, and proposed \textit{Orb-SLAM+NW-RSBA}.}
    \resizebox{.47\textwidth}{!}{
        \begin{tabular}{cccccc}
            \hline
            \multirow{2}{*}{Seq} & \multirow{2}{*}{\begin{tabular}[c]{@{}c@{}}Duration\\ {[}s{]}\end{tabular}} & \multirow{2}{*}{\begin{tabular}[c]{@{}c@{}}length\\ {[}m{]}\end{tabular}} & \multicolumn{3}{c}{Realtime factor $\epsilon$ $\uparrow$ $\mid$ Tracking duration $\underline{DUR}$ $\uparrow$}                                                                        \\ \cline{4-6}
                                 &                                             &                                             & Orb-SLAM~\cite{mur2015orb}                                                                                      & \begin{tabular}[c]{@{}c@{}}Orb-SLAM\\ +NM-RSBA~\cite{Albl2016}\end{tabular}   & \begin{tabular}[c]{@{}c@{}}Orb-SLAM\\ +NW-RSBA\end{tabular}            \\ \hline
            \#1                  & 40                                          & 46                                          & 1.47 $\mid$ \underline{0.50}                                                                                    & 1.48 $\mid$ \underline{0.28} & 1.38 $\mid$ \underline{\textbf1}      \\
            \#2                  & 27                                          & 37                                          & 1.51 $\mid$ \underline{0.90}                                                                                    & 1.40 $\mid$ \underline{0.81} & 1.40 $\mid$ \underline{\textbf1}      \\
            \#3                  & 50                                          & 44                                          & 1.47 $\mid$ \underline{0.58}                                                                                    & 1.41 $\mid$ \underline{0.36} & 1.39 $\mid$ \underline{\textbf1}      \\
            \#4                  & 38                                          & 30                                          & 1.61 $\mid$ \underline{1}                                                                                       & 1.35 $\mid$ \underline{1}    & 1.56 $\mid$ \underline{1}             \\
            \#5                  & 85                                          & 57                                          & 1.51 $\mid$ \underline{1}                                                                                       & 1.28 $\mid$ \underline{1}    & 1.38 $\mid$ \underline{1}             \\
            \#6                  & 43                                          & 51                                          & 1.47 $\mid$ \underline{0.76}                                                                                    & 1.37 $\mid$ \underline{0.76} & 1.38 $\mid$ \underline{\textbf1}      \\
            \#7                  & 39                                          & 45                                          & 1.61 $\mid$ \underline{0.89}                                                                                    & 1.47 $\mid$ \underline{0.97} & 1.49 $\mid$ \underline{\textbf1}      \\
            \#8                  & 53                                          & 46                                          & 1.56 $\mid$ \underline{0.79}                                                                                    & 1.37 $\mid$ \underline{0.96} & 1.35 $\mid$ \underline{\textbf1}      \\
            \#9                  & 45                                          & 46                                          & 1.61 $\mid$ \underline{0.14}                                                                                    & 1.51 $\mid$ \underline{0.23} & 1.55 $\mid$ \underline{\textbf{0.42}} \\
            \#10                 & 54                                          & 41                                          & 1.56 $\mid$ \underline{0.29}                                                                                    & 1.46 $\mid$ \underline{0.29} & 1.47 $\mid$ \underline{\textbf1}      \\
            t1-fast              & 28                                          & 50                                          & 1.92 $\mid$ \underline{1}                                                                                       & 1.51 $\mid$ \underline{1}    & 1.81 $\mid$ \underline{1}             \\
            t2-fast              & 29                                          & 53                                          & 1.92 $\mid$ \underline{1}                                                                                       & 1.40 $\mid$ \underline{1}    & 1.67 $\mid$ \underline{1}             \\ \hline
        \end{tabular}}
    \label{tab:realtime_factor}
\end{table}
\begin{table*}[]
    \centering
    \caption{Absolute trajectory error (ATE) of different RSBA methods after $\text{Sim}(3)$ alignment to ground truth. The best results are shown in \textcolor{green}{green}. Since some methods will lose tracking without processing the whole sequence, thus we highlight the background of each cell with different colours depending on its corresponding \textit{\underline{DUR}} value. Specifically, $\textit{\underline{DUR}}>0.9$, $0.5<\textit{\underline{DUR}}\leqslant 0.9$ and $\textit{\underline{DUR}}\leqslant0.5$ are highlighted in \colorbox[HTML]{EEFFBB}{light green}, \colorbox[HTML]{7FB3D5}{cyan}, and \colorbox[HTML]{E59866}{orange}.}
    \label{tab:APE}
    \resizebox{1\textwidth}{!}{
        \begin{tabular}{@{}cclccccccccccccc@{}}
            \toprule
            \multirow{3}{*}{Input}                                   &
            \multirow{3}{*}{Methods}                                 &
                                                                     &
            \multicolumn{13}{c}{ATE$\downarrow$}                                                                                                                                                                                                                                                                                                                                                                                                                                                                                                      \\ \cmidrule(l){3-16}
                                                                     &
                                                                     &
                                                                     &
            \multicolumn{10}{c}{TUM-RSVI~\cite{schubert2019rolling}} &
                                                                     &
            \multicolumn{2}{c}{WHU-RSVI~\cite{cao2020whu}}                                                                                                                                                                                                                                                                                                                                                                                                                                                                                            \\ \cmidrule(lr){4-13} \cmidrule(l){15-16}
                                                                     &
                                                                     &
                                                                     &
            \#1                                                      & \#2                        & \#3 & \#4                             & \#5                             & \#6                             & \#7                             & \#8                             & \#9                             & \#10                                               &                                 & t1-fast                         & t2-fast                                                                                                \\ \midrule
            \multicolumn{1}{c|}{GS data}                             &
            Orb-SLAM~\cite{mur2015orb}                               &
                                                                     &
            \cellcolor[HTML]{EEFFBB}{0.015}                          &
            \cellcolor[HTML]{EEFFBB}{0.013}                          &
            \cellcolor[HTML]{EEFFBB}{0.018}                          &
            \cellcolor[HTML]{EEFFBB}{0.107}                          &
            \cellcolor[HTML]{EEFFBB}{0.030}                          &
            \cellcolor[HTML]{EEFFBB}{0.013}                          &
            \cellcolor[HTML]{EEFFBB}{0.054}                          &
            \cellcolor[HTML]{EEFFBB}{0.053}                          &
            \cellcolor[HTML]{EEFFBB}{0.020}                          &
            \cellcolor[HTML]{EEFFBB}{0.024}                          &
                                                                     &
            \cellcolor[HTML]{EEFFBB}{0.044}                          &
            \cellcolor[HTML]{EEFFBB}{0.008}                                                                                                                                                                                                                                                                                                                                                                                                                                                                                                           \\ \midrule
            \multicolumn{1}{c|}{\multirow{3}{*}{RS data}}            & Orb-SLAM~\cite{mur2015orb} &     & \cellcolor[HTML]{E59866}{0.059} & \cellcolor[HTML]{7FB3D5}{0.100} & \cellcolor[HTML]{7FB3D5}{0.411} & \cellcolor[HTML]{EEFFBB}{0.126} & \cellcolor[HTML]{EEFFBB}{0.055} & \cellcolor[HTML]{7FB3D5}{0.044} & \cellcolor[HTML]{7FB3D5}{\textcolor{green}{0.217}} & \cellcolor[HTML]{7FB3D5}{0.218} & \cellcolor[HTML]{E59866}{0.176} & \cellcolor[HTML]{E59866}{0.373} &  & \cellcolor[HTML]{EEFFBB}{0.237} & \cellcolor[HTML]{EEFFBB}{0.018} \\
            \multicolumn{1}{c|}{}                                    &
            Orb-SLAM+NM-RSBA~\cite{Albl2016}                         &
                                                                     &
            \cellcolor[HTML]{E59866}{0.115}                          &
            \cellcolor[HTML]{7FB3D5}{0.088}                          &
            \cellcolor[HTML]{E59866}{0.348}                          &
            \cellcolor[HTML]{EEFFBB}{0.120}                          &
            \cellcolor[HTML]{EEFFBB}{0.062}                          &
            \cellcolor[HTML]{7FB3D5}{0.060}                          &
            \cellcolor[HTML]{EEFFBB}{0.251}                          &
            \cellcolor[HTML]{EEFFBB}{0.246}                          &
            \cellcolor[HTML]{E59866}{0.156}                          &
            \cellcolor[HTML]{E59866}{0.307}                          &
                                                                     &
            \cellcolor[HTML]{EEFFBB}{0.204}                          &
            \cellcolor[HTML]{EEFFBB}{0.030}                                                                                                                                                                                                                                                                                                                                                                                                                                                                                                           \\
            \multicolumn{1}{c|}{}                                    &
            Orb-SLAM+NW-RSBA (ours)                                  &
                                                                     &
            \cellcolor[HTML]{EEFFBB}{\textcolor{green}{0.011}}       &
            \cellcolor[HTML]{EEFFBB}{\textcolor{green}{0.008}}       &
            \cellcolor[HTML]{EEFFBB}{\textcolor{green}{0.031}}       &
            \cellcolor[HTML]{EEFFBB}{\textcolor{green}{0.071}}       &
            \cellcolor[HTML]{EEFFBB}{\textcolor{green}{0.034}}       &
            \cellcolor[HTML]{EEFFBB}{\textcolor{green}{0.008}}       &
            \cellcolor[HTML]{EEFFBB}{0.260}                          &
            \cellcolor[HTML]{EEFFBB}{\textcolor{green}{0.115}}       &
            \cellcolor[HTML]{E59866}{\textcolor{green}{0.028}}       &
            \cellcolor[HTML]{EEFFBB}{\textcolor{green}{0.108}}       &
                                                                     &
            \cellcolor[HTML]{EEFFBB}{\textcolor{green}{0.054}}       &
            \cellcolor[HTML]{EEFFBB}{\textcolor{green}{0.012}}                                                                                                                                                                                                                                                                                                                                                                                                                                                                                        \\ \bottomrule
        \end{tabular}}
\end{table*}
\vspace{-1.5ex}
\subsubsection{RSSLAM}
We compare the performance of conventional GS-based \textit{Orb-SLAM}~\cite{mur2015orb} versus augmented versions with \textit{NM-RSBA}~\cite{Albl2016} and proposed \textit{NW-RSBA} on 12 RS sequences.

\noindent \textbf{Real-time factor and tracking duration.} Tab.~\ref{tab:realtime_factor}
shows the statistics about 12 RS sequences and the performance of three approaches run on an AMD Ryzen 7 CPU. The results verify that all three methods achieve real-time performance. One can also confirm that the proposed \textit{Orb-SLAM+NW-RSBA} is slower than  \textit{Orb-SLAM} by a factor roughly around $1.2$  but is slightly faster than \textit{Orb-SLAM+NM-RSBA}.
As for tracking duration,  \textit{Orb-SLAM} and \textit{Orb-SLAM+NM-RSBA}~\cite{Albl2016} fail with an average $\underline{DUR}<0.5$ in most sequences once the camera moves aggressively enough. In contrast, the proposed \textit{Orb-SLAM+NW-RSBA} achieves completed tracking with $\underline{DUR}=1$ in almost all sequences.

\noindent \textbf{Absolute trajectory error.} The \textbf{ATE} results on WHU-RSVI and TUM-RSVI datasets demonstrate that the proposed \textit{Orb-SLAM+NW-RSBA} is superior
to \textit{Orb-SLAM} and \textit{Orb-SLAM+NM-RSBA} when dealing with RS effect. Qualitatively, this is clearly visible in Figs.~\ref{fig:intro} and~\ref{fig:real_slam_traj}. The sparse 3D reconstructions look much cleaner for \textit{Orb-SLAM+NW-RSBA} and close to the ground truth. The quantitative difference also becomes apparent in Tab.~\ref{tab:APE}.  \textit{Orb-SLAM+NW-RSBA} outperforms \textit{Orb-SLAM} and \textit{Orb-SLAM+NM-RSBA} both in terms of accuracy and stability.
\begin{figure*}
    \centering
    \includegraphics[width=.85\linewidth]{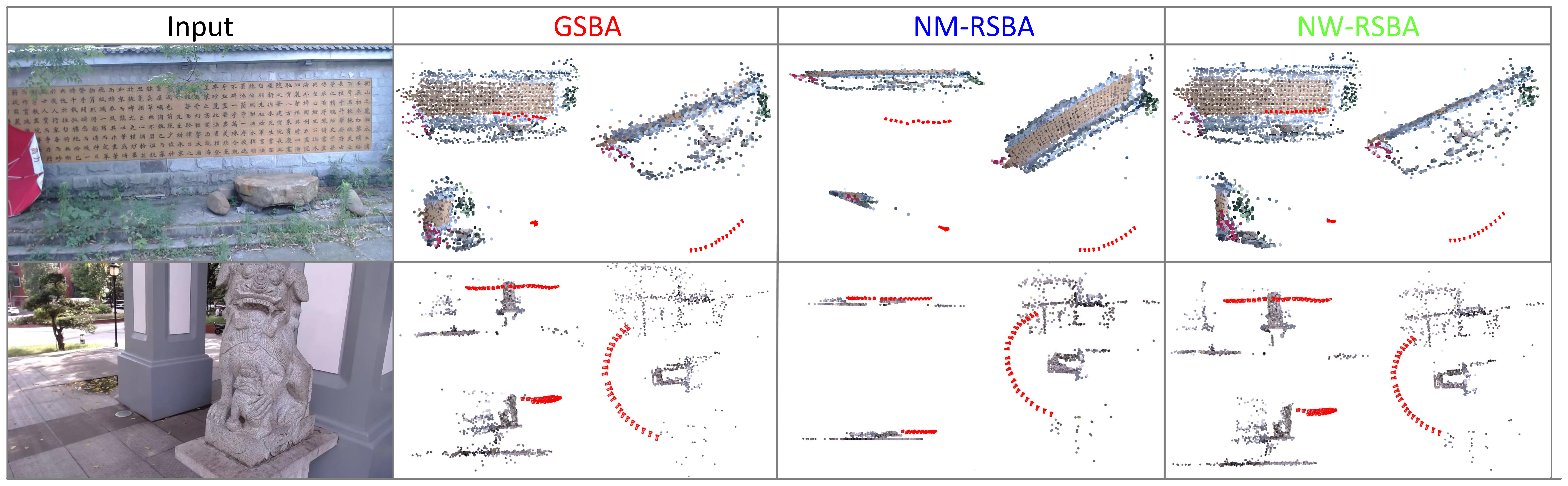}
    \caption{Three-view graph of reconstructions using SfM pipeline with \textit{GSBA}~\cite{lourakis2009sba}, \textit{NM-RSBA}~\cite{Albl2016} and proposed \textit{NW-RSBA}. }
    \label{fig:sfm}
\end{figure*}
\begin{table}[]
    \centering

    \caption{Quantitative ablation study of RSSfM on  TUM-RSVI~\cite{schubert2019rolling} and  WHU-RSVI~\cite{cao2020whu} datasets. ATE: absolute trajectory error of estimated camera pose in meters ($m$), Runtime: time cost in seconds ($s$). Best and second best results are shown in \textcolor{green}{green} and \textcolor{blue}{blue} respectively.}
    \resizebox{.47\textwidth}{!}{
        \begin{tabular}{llcc}
            \hline
            \multicolumn{1}{c}{\textbf{Ablation}} & \multicolumn{1}{c}{\textbf{Approach}} & \begin{tabular}[c]{@{}c@{}}\textbf{ATE}\\ ($m$)\end{tabular}                & \begin{tabular}[c]{@{}c@{}}\textbf{Runtime}\\ ($s$)\end{tabular} \\ \hline
            & GSBA~\cite{lourakis2009sba}           & 0.210                                     & \textcolor{green}{2.9}     \\ \hline
            & DC-RSBA~\cite{lao2018robustified}     & 0.016                                     & 1302                       \\ \hline
            No normalization \& weighting         & DM-RSBA~\cite{duchamp2015rolling}     & 0.023                                     & 740                        \\
            No weighting                          & NM-RSBA~\cite{Albl2016}               & 0.020                                     & 15.8                       \\
            No normalization                      & W-RSBA                                & \textcolor{blue}{0.013}                   & 16.1                       \\ \hline
            No Schur complement                   & NW-RSBA-0S                            & \multirow{3}{*}{\textcolor{green}{0.007}} & 16.9                       \\
            with  1-stage Schur complement        & NW-RSBA-1S                            &                                           & 13.0                       \\
            Consolidated                          & NW-RSBA-2S                            &                                           & \textcolor{blue}{9.8}      \\ \hline
        \end{tabular}}
    \label{tab:ablation}
\end{table}
\vspace{-1.5ex}
\subsubsection{RSSfM}
\noindent\textbf{Quantitative Ablation Study.}
We randomly choose 8 frames from each of the 12  RS sequences to generate 12 unordered SfM datasets and evaluate the RSBA performance via average ATE and runtime. Besides, we ablate the \textit{NW-RSBA} and compare quantitatively with related approaches. The results are presented in Tab.~\ref{tab:ablation}.
The baseline methods' performance show that \textit{NW-RSBA} obtains ATE of 0.007, which is half of the second best method \textit{DC-RSBA} and nearly $3\%$ of \textit{GSBA}. The removal of normalization from proposed \textit{NW-RSBA} adversely increases ATE up to $100\%$. The removal of covariance weighting from  \textit{NW-RSBA} adversely impacts the camera pose estimation quality with ATE growth from 0.007 to 0.020. We believe that covariance weighting helps BA leverage the RS effect and random image noise better. We ablate the consolidated  \textit{NW-RSBA-2S} to \textit{NW-RSBA-0S} by removing the proposed 2-stage Schur complement strategy and compare to \textit{NW-RSBA-1S}. The increases from 9.8s to 13.0s and 16.9s is observed for average runtime. Despite the fact \textit{NW-RSBA-2S}  is slower than \textit{GSBA} by a factor of 3, but still 2 times faster than \textit{NM-RSBA}, 2 orders of magnitude faster than \textit{DC-RSBA} and \textit{DM-RSBA}.

\noindent \textbf{Qualitative Samples.} We captured two datasets using a smartphone camera and kept the same readout direction, which is a degeneracy configuration in RSBA and will quickly lead to a planar degenerated solution for \textit{NM-RSBA}~\cite{Albl2016}. As shown in Fig.~\ref{fig:sfm} that \textit{NW-RSBA} works better in motion and 3D scene estimation while \textit{GSBA}~\cite{lourakis2009sba}  obtains a deformed reconstruction. Specifically, the sparse 3D reconstructions and recovered trajectories of \textit{NW-RSBA} look much cleaner and smoother than the ones from \textit{GSBA}. \textit{NM-RSBA} reconstructs 3D scenes which collapse into a plane since the datasets contain only one readout direction. In contrast, \textit{NW-RSBA} provides correct reconstructions. The results also verify our discussion in section~\ref{section:NW-RSBA} that error covariance weighting can handle the planar degeneracy.
\section{Conclusion}
This paper presents a novel RSBA solution without any assumption on camera image manner and type of video input. We explain the importance of conducting normalization and present a weighting technique in RSBA, which leads to normalized weighted RSBA. Extensive experiments in real and synthetic data verify the effectiveness and efficiency of the proposed \textit{NW-RSBA} method.\\

\noindent \textbf{Acknowledgements.} This work is supported by the National Key R\&D Program of China (No. 2022ZD0119003), Nature Science Foundation of China (No. 62102145), and Jiangxi Provincial 03 Special Foundation and 5G Program (Grant No. 20224ABC03A05).

{\small
    \bibliographystyle{ieee_fullname}
    \bibliography{reference}
}

\appendix
\section{Proof of Reprojection Error Covariance}
\label{supp:covariance}
\blfootnote{$\ast$ Authors contributed equally}
\blfootnote{$\dagger$ Corresponding author: \href{mailto:yizhenlao@hnu.edu.cn}{yizhenlao@hnu.edu.cn}}
\blfootnote{Project page: \href{https://delinqu.github.io/NW-RSBA}{https://delinqu.github.io/NW-RSBA}}

\indent In this section, we perform a detailed proof of reprojection error covariance. Firstly, we decompose a normalized image measurement point $\mathbf{q}_i^j = \begin{bmatrix}
c & r
\end{bmatrix}^\top$ into a perfect normalized image measurement point $\tilde{\mathbf{q}}_i^j =  \begin{bmatrix}
\tilde{c} & \tilde{r}
\end{bmatrix}^\top$ and a normalized image Gaussian  measurement noise $\begin{bmatrix}
n_c &n_r \end{bmatrix}^\top$:
\begin{align}
&\begin{bmatrix}
c & r
\end{bmatrix}^\top = \begin{bmatrix}
\tilde{c} & \tilde{r}
\end{bmatrix}^\top + \begin{bmatrix}
n_c & n_r
\end{bmatrix}^\top,
\label{equation:noise_modling}
\end{align}
with,
\begin{align}
&\begin{bmatrix}
n_c & n_r
\end{bmatrix}^\top \sim \mathcal{N}(0, \mathbf{W\Sigma }\mathbf{W}^{\top}), 
\label{supp:prior_noise_definition}
\end{align}
\begin{align}
&\mathbf{W} =  \begin{bmatrix}
1/f_x & 0\\ 
0  &  1/f_y
\end{bmatrix},
\label{supp:W_definition}
\end{align}

\noindent where $\mathbf{\Sigma}$ is the prior Gaussian measurement noise and $f_x \ f_y$ are the x-axis and y-axis focal length respectively. Then we can substitute Eq.~(\ref{equation:noise_modling}) to the normalized measurement based reprojection cost function as:
\begin{equation}
    \resizebox{0.9\linewidth}{!}{
        \begin{math}
        \begin{aligned}
        {\mathbf{e}}_{i}^{j} &= \begin{bmatrix}
        {c}_{i}^{j} & {r}_{i}^{j}
        \end{bmatrix}^\top - \Pi (
        \mathbf{R}^j(r_{i}^{j})\mathbf{P}_{i} +  \mathbf{t}^j(r_{i}^{j})
        ) \\
        & = \begin{bmatrix}
        {c}_{i}^{j} & {r}_{i}^{j}
        \end{bmatrix}^\top - \Pi (
        \mathbf{R}^j(\tilde{r}_i^j + n_r)\mathbf{P}_{i} +  \mathbf{t}^j(\tilde{r}_i^j + n_r)
        ),
        \end{aligned} 
        \end{math}
        }
\label{equation:proof1}
\end{equation}  
\noindent where $\Pi (\mathbf{R}^j(\tilde{r}_i^j + n_r)\mathbf{P}_{i} +  \mathbf{t}^j(\tilde{r}_i^j + n_r))$ can be linearized using the Taylor first order approximation:  
\begin{equation}
\begin{aligned}
&\Pi (\mathbf{R}^j(\tilde{r}_i^j + n_r)\mathbf{P}_{i} +  \mathbf{t}^j(\tilde{r}_i^j + n_r))
\\ \approx &\begin{bmatrix}
\tilde{c}_{i}^{j}\\ 
\tilde{r}_{i}^{j}
\end{bmatrix} + \frac{\partial \Pi (\mathbf{R}^j({r}_i^j)\mathbf{P}_{i} +  \mathbf{t}^j({r}_i^j))}{\partial n_r}n_r.
\end{aligned} 
\label{equation:proof2}
\end{equation}
\noindent Then by substituting Eq.~(\ref{equation:proof2}) into  Eq.~(\ref{equation:proof1}), we have
\begin{equation}
    \resizebox{0.90\linewidth}{!}{
        \begin{math}
        \begin{aligned}
        \mathbf{{e}}_i^j = \begin{bmatrix}
        n_c\\ 
        n_r
        \end{bmatrix} - \frac{\partial \Pi (\mathbf{R}^j({r}_i^j)\mathbf{P}_{i} + \mathbf{t}^j({r}_i^j))}{\partial n_r}n_r 
        = \mathbf{C}_i^j\begin{bmatrix}
        n_c\\ 
        n_r
        \end{bmatrix}.
        \end{aligned} 
    \end{math}
    }
\label{equation:proof3}
\end{equation}
By applying the chain rule of derivation, we can get the analytical formulation of matrix $\mathbf{C}_i^j$.
\begin{equation}
    \resizebox{0.90\linewidth}{!}{
        \begin{math}
        \begin{aligned}
        \frac{\partial \Pi (\mathbf{R}^j({r}_i^j)\mathbf{P}_{i} + \mathbf{t}^j({r}_i^j))}{\partial n_r} = \frac{\partial \Pi (\mathbf{R}^j({r}_i^j)\mathbf{P}_{i} + \mathbf{t}^j({r}_i^j))}{\partial {\mathbf{P}^c}_i^j} \frac{\partial {\mathbf{P}^c}_i^j}{\partial n_r}, 
        \end{aligned} 
        \end{math}
        }
\label{supp:chain_de}
\end{equation}
with,
\begin{equation}
\begin{aligned}
\frac{\partial \Pi (\mathbf{R}^j({r}_i^j)\mathbf{P}_{i} + \mathbf{t}^j({r}_i^j))}{\partial {\mathbf{P}^c}_i^j} = {\begin{bmatrix}
\frac{1}{{{Z^c}_i^j}} & 0 \\ 
0 &  \frac{1}{{{Z^c}_i^j}} \\ 
\frac{-{{X^c}_i^j}}{{{Z^c}_i^j}^2} & \frac{-{{Y^c}_i^j}}{{{Z^c}_i^j}^2}
\end{bmatrix}}^{\top},
\end{aligned} 
\label{supp:chain_de_supp1}
\end{equation}
\begin{equation}
\begin{aligned}
\frac{\partial {\mathbf{P}^c}_i^j}{\partial n_r} = [\boldsymbol{\omega}^j]_\times\mathbf{R}^j\mathbf{P}_i+\mathbf{d}^j.
\end{aligned} 
\label{supp:chain_de_supp2}
\end{equation}
By substituting Eq.~(\ref{supp:chain_de}) into Eq.~(\ref{equation:proof3}), we can get analytical formulation:
\begin{equation}\label{supp:sigma_c} 
\resizebox{0.85\linewidth}{!}{
    \begin{math}
    \begin{aligned}
    \mathbf{C}_i^j = \begin{bmatrix}1 &0 \\ 0 &1\end{bmatrix} - \begin{bmatrix}
    \frac{1}{{{Z^c}_i^j}} & 0 \\ 
    0 &  \frac{1}{{{Z^c}_i^j}} \\ 
    \frac{-{{X^c}_i^j}}{{{Z^c}_i^j}^2} & \frac{-{{Y^c}_i^j}}{{{Z^c}_i^j}^2}
    \end{bmatrix}^\top ([\boldsymbol{\omega}^j]_\times\mathbf{R}^j\mathbf{P}_i+\mathbf{d}^j) {\begin{bmatrix}0 \\1 \end{bmatrix}}^\top,
    \end{aligned}
    \end{math}
    }
\end{equation}
\noindent where ${\mathbf{P}^c}_i^j = {\begin{bmatrix}{X^c}_i^j &{Y^c}_i^j&{Z^c}_i^j\end{bmatrix}}^\top = \mathbf{R}^j({r}_i^j)\mathbf{P}_{i} + \mathbf{t}^j({r}_i^j)$ is the world point $\mathbf{P}_{i}$ in camera $j$ coordinates. Combining Eq.~(\ref{supp:prior_noise_definition}) with Eq.~(\ref{equation:proof3}), we prove that $\mathbf{{e}}_i^j$ follows a weighted Gaussian distribution:
\begin{equation}
\begin{aligned}
\mathbf{{e}}_i^j \sim \mathcal{N} ( 0, \mathbf{C}_i^j \mathbf{W} \mathbf{\Sigma}\mathbf{W}^{\top} {\mathbf{C}_i^j}^\top ) .
\end{aligned} 
\label{equation:proofr}
\end{equation} 
\\
\section{Analytical Jacobian matrix Derivation}
\label{supp:Jacobian}
In this section, we provide a detailed derivation of  the analytical Jacobian matrix used in our proposed \textit{NW-RSBA} solution. 
\subsection{Jacobian Matrix Parameterization}
\label{section:jacobian_parameterization}
\indent To derive the analytical Jacobian matrix of Eq.~(\ref{supp:supp_J_matrix}), we use  ${\boldsymbol{\xi}_i^j}\in \mathbf{so}({3})$ to parametrize ${\mathbf R_i^j} \in \mathbf{SO}({3})$. These two representations can be transformed to each other by Rodrigues formulation ${\mathbf R_i^j} = \mathbf{Exp}(\boldsymbol{\xi}_i^j)$ and ${\boldsymbol {\xi}_i^j} = \mathbf{Log}(\mathbf{R}_i^j)$, which are defined as:

\begin{equation}
\begin{aligned}
{\mathbf R} &= \mathbf{Exp}(\boldsymbol{\xi}) \\
&= \mathbf{I} + \frac{sin(\left \| \boldsymbol{\xi} \right \|)}{\left \| \boldsymbol{\xi} \right \|} \boldsymbol{\xi}^{\wedge } + \frac{1 - cos(\left \| \boldsymbol{\xi} \right \|)}{{\left \| \boldsymbol{\xi} \right \|}^2} {(\boldsymbol{\xi}^{\wedge })}^2.
\end{aligned} 
\label{supp:exp}
\end{equation}  

\begin{equation}
\begin{aligned}
{\boldsymbol \xi} &= \mathbf{Log}(\mathbf{R}) \\
&= \frac{\boldsymbol\theta}{2sin(\boldsymbol\theta)}{(\mathbf {R} - \mathbf{R}^{\top})}^{\vee},
\end{aligned} 
\label{supp:log}
\end{equation}  
\noindent with,
\begin{align}
\boldsymbol\theta = arccos((tr(\mathbf R) - 1)/2),
\end{align} 
where ${\wedge}$ is the skew-symmetric operator that can transform a vector to the corresponding skew-symmetric matrix. Besides, ${\vee}$ is the inverse operator.

\subsection{Partial Derivative of Reprojection error}
\label{section:partial_derivative_error}
Recall the normalized weighted error term, which is defined as:
\begin{align}
{{\mathbf{\hat{e}}}_i^j } = \mathbf{\Sigma}^{-\frac{1}{2}}{{\mathbf{W}}^{-1}{{\mathbf{C}_i^j}}^{-1}\mathbf e_i^j}.
\label{supp:supp_J_matrix}
\end{align}
Then we can get five atomic partial derivatives of $\partial{{\mathbf{\hat{e}}_i^j }}$ over $\partial{{\mathbf{P}_{i}}}$, $\partial\boldsymbol{\xi}^{j}$, ${\partial \mathbf{t}^{j}}$, ${\partial \boldsymbol{\omega}^{j}}$ and ${\partial \mathbf{d}^{j}}$ as: 

\begin{equation}
\resizebox{0.85\linewidth}{!}{
\begin{math}
\begin{aligned}
\frac {\partial{{\mathbf{\hat{e}}_i^j}}}{\partial{\mathbf{P}_{i}}}={{\mathbf{\Sigma}_{n}}^{-\frac{1}{2}} \mathbf{W}^{-1} }({{\mathbf {C}_{i}^j}^{-1}} \frac {\partial {\mathbf{e}_{i}^j}}{\partial{\mathbf{P}^c}_i^j} \frac {\partial{\mathbf{P}^c}_i^j}{\partial\mathbf{P}_{i}} +  \begin{bmatrix}{{\mathbf{e}_i^j}^\top\frac {\partial{{{\mathbf C_{i}^j}}(1)^{-\top}}}{\partial\mathbf{P}_{i}}} \\{\mathbf{e}_i^j}^\top\frac {\partial{{{\mathbf C_{i}^j}}(2)^{-\top}}}{\partial\mathbf{P}_{i}}\end{bmatrix}),
\label{supp:PD_pi}
\end{aligned}
\end{math}
}
\end{equation}

\begin{equation}
\resizebox{0.85\linewidth}{!}{
\begin{math}
\begin{aligned}
\frac {\partial{\mathbf{\hat{e}}_i^j}}{\partial\boldsymbol{\xi}^{j}}={{\mathbf{\Sigma}_{n}}^{-\frac{1}{2}} \mathbf{W}^{-1} }({{\mathbf C_{i}^j}}^{-1} \frac {\partial {\mathbf e_{i}^j}}{\partial {\mathbf{P}^c}_i^j} \frac {\partial{\mathbf{P}^c}_i^j}{\partial\boldsymbol{\xi}^{j}} +  \begin{bmatrix}{{\mathbf e_i^j}}^\top\frac {\partial{{\mathbf C_{i}^j}(1)^{-\top}}}{\partial \mathbf{\boldsymbol\xi}^{j}} \\{{\mathbf e_i^j}}^\top\frac {\partial{{\mathbf C_{i}^j}(2)^{-\top}}}{\partial \boldsymbol{\xi}^{j}}\end{bmatrix}),
\label{supp:PD_ri}
\end{aligned}
\end{math}
}
\end{equation}

\begin{equation}
\resizebox{0.85\linewidth}{!}{
\begin{math}
\begin{aligned}
\frac {\partial {\mathbf{\hat{e}}_i^j}}{\partial \mathbf{t}^{j}}={{\mathbf{\Sigma}_{n}}^{-\frac{1}{2}} \mathbf{W}^{-1} }({\mathbf C_{i}^j}^{-1} \frac {\partial{\mathbf e_{i}^j}}{\partial {\mathbf{P}^c}_i^j} \frac {\partial {\mathbf{P}^c}_i^j}{\partial \mathbf{t}^{j}} +  \begin{bmatrix}{{{\mathbf e_i^j}}^\top\frac {\partial{{\mathbf C_{i}^j}(1)^{-\top}}}{\partial \mathbf{t}^{j}}} \\{\mathbf{e}_i^j}^\top\frac {\partial{{\mathbf C_{i}^j}(2)^{-\top}}}{\partial \mathbf{t}^{j}}\end{bmatrix}), 
\label{supp:PD_ti}
\end{aligned}
\end{math}
}
\end{equation}

\begin{equation}
\resizebox{0.85\linewidth}{!}{
\begin{math}
\begin{aligned}
\frac {\partial {\mathbf{\hat{e}}_i^j}}{\partial \boldsymbol{\omega}^{j}}={{\mathbf{\Sigma}_{n}}^{-\frac{1}{2}} \mathbf{W}^{-1} }({ \mathbf C_{i}^j}^{-1} \frac {\partial{ \mathbf e_{i}^j}}{\partial {\mathbf{P}^c}_i^j} \frac {\partial {\mathbf{P}^c}_i^j}{\partial \boldsymbol{\omega}^{j}} +  \begin{bmatrix}{{\mathbf{e}_i^j}^\top\frac {\partial{{\mathbf C_{i}^j}(1)^{-\top}}}{\partial \boldsymbol{\omega}^{j}}} \\{\mathbf{e}_i^j}^\top\frac {\partial{{\mathbf C_{i}^j}(2)^{-\top}}}{\partial \boldsymbol{\omega}^{j}}\end{bmatrix}),
\label{supp:PD_wi} 
\end{aligned}
\end{math}
}
\end{equation}

\begin{equation}
\resizebox{0.85\linewidth}{!}{
\begin{math}
\begin{aligned}
\frac {\partial {\mathbf{\hat{e}}_i^j}}{\partial \mathbf{d}^{j}}={{\mathbf{\Sigma}_{n}}^{-\frac{1}{2}} \mathbf{W}^{-1} }({\mathbf C_{i}^j}^{-1} \frac {\partial{\mathbf e_{i}^j}}{\partial{\mathbf{P}^c}_i^j} \frac {\partial{\mathbf{P}^c}_i^j}{\partial\mathbf{d}^{j}} +  \begin{bmatrix}{{\mathbf{e}_i^j}^\top\frac {\partial{{\mathbf C_{i}^j}(1)^{-\top}}}{\partial \mathbf{d}^{j}}} \\{\mathbf{e}_i^j}^\top\frac {\partial{{\mathbf C_{i}^j}(2)^{-\top}}}{\partial \mathbf{d}^{j}}\end{bmatrix}),
\label{supp:PD_di}
\end{aligned}
\end{math}
} 
\end{equation}

with,

\begin{align}
\frac {\partial{\mathbf e_{i}^{j}}}{\partial{\mathbf{P}^c}_i^j} &= -
\begin{bmatrix}
\frac{1}{{{Z}^{c}}_i^j}& 0 & - \frac{{{X}^{c}}_i^j}{({{Z}^{c}}_i^j)^{2}}\\
0 & \frac{1}{{{Z}^{c}}_i^j} & - \frac{{{Y}^{c}}_i^j}{({{Z}^{c}}_i^j)^{2}}\\
\end{bmatrix}, \\
\frac {\partial{\mathbf{P}^c}_i^j}{\partial\mathbf{P}_{i}} &= (\mathbf{I}+[\boldsymbol{\omega}^j]_{\times} {{r}_{i}^j}){\mathbf R^j},\\
\frac {\partial{\mathbf{P}^c}_i^j}{\partial\boldsymbol{\xi}^j} &= - (\mathbf{I}+[\boldsymbol{\omega}^j]_{\times} {r}_{i}^j)[{\mathbf R^j}\mathbf{P}_{i}]_{\times}, \\
\frac {\partial{\mathbf{P}^c}_i^j}{\partial{\mathbf t^j}} &= [\mathbf{I}]_{3\times3}, \\
\frac {\partial{\mathbf{P}^c}_i^j}{\partial\boldsymbol{\omega}^j} &= - r_i^j[{\mathbf R^j}\mathbf{P}_i]_{\times}, \\
\frac {\partial{\mathbf{P}^c}_i^j}{\partial{\mathbf d^j}} &= {r}_{i}^j[\mathbf{I}]_{3\times3},
\end{align}
where ${\mathbf C_{i}^j}(1)$ and ${\mathbf C_{i}^j}(2)$ represents the first and second row of matrix ${\mathbf C_{i}^j}$ respectively. 
\\
\\
\indent We further need to derive the partial derivatives of $\partial{{\mathbf C_{i}^j}(.)^{-\top}}$ over $\partial{{\mathbf{P}_{i}}}$, $\partial\boldsymbol{\xi}^{j}$, ${\partial \mathbf{t}^{j}}$, ${\partial \boldsymbol{\omega}^{j}}$ and ${\partial \mathbf{d}^{j}}$ in Eq.~(\ref{supp:PD_pi} - \ref{supp:PD_di}). Recall the $\mathbf{C}_i^j$ and $\mathbf{W} $ definition in Eq.~(\ref{supp:sigma_c}) and Eq.~(\ref{supp:W_definition}). For convenience, we define the following two intermediate variables:
\begin{align}
\boldsymbol\gamma_i^j &={\begin{bmatrix}
\frac{1}{{{Z^c}_i^j}} & 0 \\ 
0 &  \frac{1}{{{Z^c}_i^j}} \\ 
\frac{-{{X^c}_i^j}}{{{Z^c}_i^j}^2} & \frac{-{{Y^c}_i^j}}{{{Z^c}_i^j}^2}
\end{bmatrix}}^{\top},   \label{supp:gamma} \\
\boldsymbol\delta_i^j &=  [\boldsymbol{\omega}^j]_{\times} {\mathbf R^j}\mathbf{P}_i + \mathbf d^j.   \label{supp:delta}
\end{align}
\noindent Then we can rewrite Eq.~(\ref{supp:sigma_c}) as:
\begin{align}
	\mathbf C_i^j = \begin{bmatrix}1 &0 \\ 0 &1\end{bmatrix} -\boldsymbol\gamma_i^j \boldsymbol\delta_i^j \begin{bmatrix}0 &1 \end{bmatrix}=\begin{bmatrix}1 &-\boldsymbol\gamma_i^j \boldsymbol\delta_i^j \\ 0 &1 - \boldsymbol\gamma_i^j \boldsymbol\delta_i^j\end{bmatrix}, \label{supp:C_definition_new}
\end{align}
\noindent and its inverse formulation as:
\begin{align}
{\mathbf C_i^j}^{-1} =\begin{bmatrix}1 &\frac{\boldsymbol\gamma_i^j \boldsymbol\delta_i^j}{1 - \boldsymbol\gamma_i^j \boldsymbol\delta_i^j} \\ 0 & \frac{1}{1 - \boldsymbol\gamma_i^j \boldsymbol\delta_i^j} \end{bmatrix}. \label{supp:C_definition_new_inverse}
\end{align}

\noindent  Then we can derive the partial derivative as:
\begin{equation}
\label{supp:pertial_1_C_P}
\frac {\partial{{\mathbf C_{i}^j}(1)^{-\top}}}{\partial \mathbf{P}_{i}}=\begin{bmatrix}[0]_{1\times3} \\\frac{(1 - \boldsymbol\beta)\frac{\partial{\boldsymbol\alpha_i^j}}{\partial{\mathbf P_i}} + \boldsymbol\alpha\frac{\partial{\boldsymbol\beta_i^j}}{\partial{\mathbf P_i}}}{(1-\boldsymbol\beta_i^j)^2} \end{bmatrix},
\end{equation}
\begin{equation}
\frac {\partial {\mathbf C_i^j(2)}^{-\top}}{\partial\mathbf{P}_{i}}=\begin{bmatrix}[0]_{1\times3} \\\frac{\frac{\partial{\boldsymbol\beta_i^j}}{\partial{\mathbf{P}_{i}}}}{(1-\boldsymbol\beta_i^j)^2} \end{bmatrix},
\end{equation}
\begin{equation}
\frac {\partial{{\mathbf C_{i}^j}(1)^{-\top}}}{\partial{\boldsymbol\xi}^{j}}=\begin{bmatrix}[0]_{1\times3} \\\frac{(1 - \boldsymbol\beta_i^j)\frac{\partial{\boldsymbol\alpha_i^j}}{\partial{\boldsymbol\xi^j}} + \boldsymbol\alpha_i^j\frac{\partial{\boldsymbol\beta_i^j}}{\partial{\boldsymbol\xi^j}}}{(1-\boldsymbol\beta_i^j)^2} \end{bmatrix}, 
\end{equation}
\begin{equation}
\frac {\partial {\mathbf C_i^j(2)}^{-\top}}{\partial{\boldsymbol\xi}^{j}}=\begin{bmatrix}[0]_{1\times3} \\\frac{\frac{\partial{\boldsymbol\beta_i^j}}{\partial{{\boldsymbol\xi}^{j}}}}{(1-\boldsymbol\beta_i^j)^2} \end{bmatrix},
\end{equation}
\begin{equation}
\frac {\partial{{\mathbf C_{i}^j}(1)^{-\top}}}{\partial\mathbf{t}^{j}}=\begin{bmatrix}[0]_{1\times3} \\\frac{(1 - \boldsymbol\beta_i^j)\frac{\partial{\boldsymbol\alpha_i^j}}{\partial{\mathbf t^j}} + \boldsymbol\alpha_i^j\frac{\partial{\boldsymbol\beta_i^j}}{\partial{\mathbf t^j}}}{(1-\boldsymbol\beta_i^j)^2} \end{bmatrix},
\end{equation}
\begin{equation}
\frac {\partial {\mathbf C_i^j(2)}^{-\top}}{\partial\mathbf{t}^{j}}=\begin{bmatrix}[0]_{1\times3} \\\frac{\frac{\partial{\boldsymbol\beta_i^j}}{\partial{\mathbf{t}^{j}}}}{(1-\boldsymbol\beta_i^j)^2} \end{bmatrix},
\end{equation}
\begin{equation}
\frac {\partial{{\mathbf C_{i}^j}(1)^{-\top}}}{\partial\boldsymbol{\omega}^{j}}=\begin{bmatrix}[0]_{1\times3} \\\frac{(1 - \boldsymbol\beta_i^j)\frac{\partial{\boldsymbol\alpha_i^j}}{\partial{\boldsymbol \omega^j}} + \boldsymbol\alpha_i^j\frac{\partial{\boldsymbol\beta_i^j}}{\partial{\boldsymbol\omega^j}}}{(1-\boldsymbol\beta_i^j)^2} \end{bmatrix},
\end{equation}
\begin{equation}
\frac {\partial {\mathbf C_i^j(2)}^{-\top}}{\partial\boldsymbol{\omega}^{j}}=\begin{bmatrix}[0]_{1\times3} \\\frac{\frac{\partial{\boldsymbol\beta_i^j}}{\partial{{\boldsymbol\omega}^{j}}}}{(1-\boldsymbol\beta_i^j)^2} \end{bmatrix},
\end{equation}
\begin{equation}
\frac {\partial{{\mathbf C_{i}^j}(1)^{-\top}}}{\partial\mathbf{d}^{j}}=\begin{bmatrix}[0]_{1\times3} \\  \frac{(1 - \boldsymbol\beta_i^j)\frac{\partial{\boldsymbol\alpha_i^j}}{\partial{\mathbf d^j}} + \boldsymbol\alpha_i^j\frac{\partial{\boldsymbol\beta_i^j}}{\partial{\mathbf d^j}}}{(1-\boldsymbol\beta_i^j)^2}   \end{bmatrix},
\end{equation}
\begin{equation}
\label{supp:pertial_2_C_d}
\frac {\partial {\mathbf C_i^j(2)}^{-\top}}{\partial\mathbf{d}^{j}}=\begin{bmatrix}[0]_{1\times3} \\ \frac{\frac{\partial{\boldsymbol\beta_i^j}}{\partial{\mathbf{d}^{j}}}}{(1-\boldsymbol\beta_i^j)^2}\end{bmatrix},
\end{equation}
\noindent where $\mathbf{C}(1)$ and  $\mathbf{C}(2)$ are the first and second row of  $\mathbf{C}$ respectively, and two intermediate variables $\boldsymbol\alpha_i^j \ \boldsymbol\beta_i^j$ are the first and second row of $\boldsymbol\gamma_i^j \boldsymbol\delta_i^j$ respectively 
\begin{align}
\boldsymbol\gamma_i^j \boldsymbol\delta_i^j  =  \begin{bmatrix}\boldsymbol\alpha_i^j\\ \boldsymbol\beta_i^j \end{bmatrix}. \label{supp:alphg_beta}
\end{align}
\\
\\
\indent Finally we have to derive the partial derivative of $\partial{\boldsymbol\alpha_i^j}$ and $\partial{\boldsymbol\beta_i^j}$  over $\partial{{\mathbf{P}_{i}}}$, $\partial\boldsymbol{\xi}^{j}$, ${\partial \mathbf{t}^{j}}$, ${\partial \boldsymbol{\omega}^{j}}$ and ${\partial \mathbf{d}^{j}}$ in Eq.~(\ref{supp:pertial_1_C_P} - \ref{supp:pertial_2_C_d}):
\begin{align}
\frac{\partial(\boldsymbol\gamma_i^j \boldsymbol\delta_i^j)}{\partial\mathbf{P}_{i}}=\boldsymbol\gamma_i^j \frac{\partial \boldsymbol\delta_i^j }{\partial {\mathbf P_i}} + \begin{bmatrix}{\boldsymbol\delta_i^j}^\top \frac{\partial\boldsymbol\gamma_i^j(1)^\top}{\partial{\mathbf{P}^c}_i^j}\frac{\partial{\mathbf{P}^c}_i^j}{\partial{\mathbf P_i}}\\
{\boldsymbol\delta_i^j}^\top \frac{\partial\boldsymbol\gamma_i^j(2)^\top}{\partial{\mathbf{P}^c}_i^j}\frac{\partial{\mathbf{P}^c}_i^j}{\partial{\mathbf P_i}}  \end{bmatrix}, \end{align}
\begin{align}
\frac{\partial(\boldsymbol\gamma_i^j \boldsymbol\delta_i^j)}{\partial{\boldsymbol\xi}^{j}}=\mathbf\gamma_i^j \frac{\partial \boldsymbol\delta_i^j }{\partial {\boldsymbol{\xi}^j}} + \begin{bmatrix}{\boldsymbol\delta_i^j}^\top \frac{\partial\boldsymbol\gamma_i^j(1)^\top}{\partial{\mathbf{P}^c}_i^j}\frac{\partial{\mathbf{P}^c}_i^j}{\partial{\boldsymbol{\xi}^j}}\\{\boldsymbol\delta_i^j}^\top \frac{\partial\boldsymbol\gamma_i^j(2)^\top}{\partial{\mathbf{P}^c}_i^j}\frac{\partial{\mathbf{P}^c}_i^j}{\partial{\boldsymbol{\xi}^j}}  \end{bmatrix}, \end{align}
\begin{align}
\frac{\partial(\boldsymbol\gamma_i^j \boldsymbol\delta_i^j)}{\partial\mathbf{t}^{j}}=\boldsymbol\gamma_i^j \frac{\partial \boldsymbol\delta_i^j }{\partial {\mathbf t^j}} + \begin{bmatrix}{\boldsymbol\delta_i^j}^\top \frac{\partial\boldsymbol\gamma_i^j(1)^\top}{\partial{\mathbf{P}^c}_i^j}\frac{\partial{\mathbf{P}^c}_i^j}{\partial{\mathbf t^j}}\\{\boldsymbol\delta_i^j}^\top \frac{\partial\boldsymbol\gamma_i^j(2)^\top}{\partial{\mathbf{P}^c}_i^j}\frac{\partial{\mathbf{P}^c}_i^j}{\partial{\mathbf t^j}}  \end{bmatrix},\end{align}
\begin{align}
\frac{\partial(\boldsymbol\gamma_i^j \boldsymbol\delta_i^j)}{\partial\boldsymbol{\omega}^{j}}=\boldsymbol\gamma_i^j \frac{\partial \boldsymbol\delta_i^j }{\partial {\boldsymbol \omega^j}} + \begin{bmatrix}{\boldsymbol\delta_i^j}^\top \frac{\partial\boldsymbol\gamma_i^j(1)^\top}{\partial{\mathbf{P}^c}_i^j}\frac{\partial{\mathbf{P}^c}_i^j}{\partial{\boldsymbol\omega^j}}\\{\boldsymbol\delta_i^j}^\top \frac{\partial\boldsymbol\gamma_i^j(2)^\top}{\partial{\mathbf{P}^c}_i^j}\frac{\partial{\mathbf{P}^c}_i^j}{\partial{\boldsymbol\omega^j}}  \end{bmatrix}, \end{align}
\begin{align}
\frac{\partial(\boldsymbol\gamma_i^j \boldsymbol\delta_i^j)}{\partial\mathbf{d}^{j}}=\boldsymbol\gamma_i^j \frac{\partial \boldsymbol\delta_i^j }{\partial {\mathbf d^j}} + \begin{bmatrix}{\boldsymbol\delta_i^j}^\top \frac{\partial\boldsymbol\gamma_i^j(1)^\top}{\partial{\mathbf{P}^c}_i^j}\frac{\partial{\mathbf{P}^c}_i^j}{\partial{\mathbf d^j}}\\{\boldsymbol\delta_i^j}^\top \frac{\partial\boldsymbol\gamma_i^j(2)^\top}{\partial{\mathbf{P}^c}_i^j}\frac{\partial{\mathbf{P}^c}_i^j}{\partial{\mathbf d^j}}  \end{bmatrix},
\end{align}
with, 
\begin{align}
\frac{\partial\boldsymbol\gamma_i^j(1)^\top}{\partial{\mathbf{P}^c}_i^j}&=\begin{bmatrix}0 &0 &-\frac{1}{{{Z^c}_i^j}^2}  \\ 0 &0 &0\\-\frac{1}{{{Z^c}_i^j}^2} &0 &\frac{2{{X^c}_i^j}}{{{Z^c}_i^j}^3} \end{bmatrix}, \end{align}
\begin{align}
\frac{\partial\boldsymbol\gamma_i^j(2)^\top}{\partial{\mathbf{P}^c}_i^j}&=\begin{bmatrix}0 &0 &0 \\ 0 &0 &-\frac{1}{{{Z^c}_i^j}^2}\\0 &-\frac{1}{{{Z^c}_i^j}^2} &\frac{2{{Y^c}_i^j}}{{{Z^c}_i^j}^3} \end{bmatrix}, \end{align}
\begin{align}
\frac{\partial \boldsymbol\delta_i^j }{\partial {\mathbf P_i}}&=[\boldsymbol{\omega}^j]_{\times}\mathbf R^j, \\
\frac{\partial \boldsymbol\delta_i^j }{\partial {\boldsymbol{\xi}^j}}&=[\boldsymbol{\omega}^j]_{\times}[{\mathbf R^j}\mathbf{P}_{i}]_{\times}, \\
\frac{\partial \boldsymbol\delta_i^j }{\partial {\mathbf t^j}}&=[0]_{3\times3},\\
\frac{\partial \boldsymbol\delta_i^j }{\partial {\boldsymbol \omega^j}}&=-[{\mathbf R^j}\mathbf P_i]_{\times},
\\
\frac{\partial \boldsymbol\delta_i^j }{\partial {\mathbf d^j}}&=[\mathbf{I}]_{3\times3}.
\end{align}
\\
\\
\section{The proof of degeneracy resilience ability}
\label{supp:degeneracy_resilience_ability}
As proved in~\cite{Albl2016}, under the planar degeneracy configuration, the y-component of the reprojection error will reduce to zero. To say it in another way, the noise perturbation along the y-component of the observation will not be reflected in the y-component of the reprojection error (remains at zero). The reprojection error covariance matrix must have a zero variance in the y-coordinate of its values according to the definition of covariance. We prove this theoretically in the following.

In correspondence with notations in the manuscript, we first define ${\mathbf{P}^g}_i^j = [{X^g}_i^j \ {Y^g}_i^j \ {Z^g}_i^j]^\top =\mathbf{R}_{0}^j \mathbf{P}_{i} +  \mathbf{t}_{0}^j$
and it can be related with ${\mathbf{P}^c}_i^j$ as $
 [\boldsymbol{\omega}^j]_\times\mathbf{R}^j_0\mathbf{P}_i+\mathbf{d}^j = ({\mathbf{P}^c}_i^j - {\mathbf{P}^g}_i^j)\ /\ {{v}_i^j}
$. Then we rewrite the Eq.~(\ref{supp:sigma_c}):

\begin{equation}\label{supp:ability_proof} 
\resizebox{0.85\linewidth}{!}{
	\begin{math}
\begin{aligned}
\mathbf{C}_i^j &= \begin{bmatrix}1 &0 \\ 0 &1\end{bmatrix} - \begin{bmatrix}
\frac{1}{{{Z^c}_i^j}} & 0 \\ 
0 &  \frac{1}{{{Z^c}_i^j}} \\ 
\frac{-{{X^c}_i^j}}{{{Z^c}_i^j}^2} & \frac{-{{Y^c}_i^j}}{{{Z^c}_i^j}^2}
\end{bmatrix}^\top\frac{{\mathbf{P}^c}_i^j - {\mathbf{P}^g}_i^j}{{v}_i^j}  {\begin{bmatrix}0 \\1 \end{bmatrix}}^\top\\
&=\begin{bmatrix}1 &0 \\ 0 &1\end{bmatrix} - \begin{bmatrix}\frac{{Z^g}_i^j {X^c}_i^j - {Z^c}_i^j {X^g}_i^j}{{v}_i^j{{Z^c}_i^j}^2} \\ \frac{{Z^g}_i^j {Y^c}_i^j - {Z^c}_i^j {Y^g}_i^j}{{v}_i^j{{Z^c}_i^j}^2}\end{bmatrix}{\begin{bmatrix}0 \\1 \end{bmatrix}}^\top.
\end{aligned}
\end{math}
}
\end{equation}

Under the degeneracy configuration, the observed point will project to the plane $y = 0$ in the camera coordinate. We then substitute the degeneracy condition ${Y^g}_i^j = 0, {Z^g}_i^j = {Z^c}_i^j, {v}_i^j = {Y^c}_i^j / {Z^c}_i^j$ into Eq.(\ref{supp:ability_proof}). It can be verified that the lower right component of $\mathbf{C}_i^j$ reduces to zero, which means that the y-coordinate variance in the reprojection error covariance matrix will reduce to zero.

Based on the explicitly modeled reprojection error covariance, we can decompose its inverse form and then reweight the reprojection error Eq.(11-13), which will result an isotropy covariance (Fig.~3). The corresponding weight will rapidly approach infinite during the degeneracy process since the y-coordinate variance gradually reduces to zero. As a result, the reweighted reprojection error will grow exponentially, which will prevent the continuation of the degeneracy. The error of \textit{NM-RSBA} decreases gradually during the degeneracy process, while \textit{NW-RSBA} grows exponentially and converges around the ground truth.

\section{The equivalent between Normalized DC-RSBA and NW-RSBA }
\label{supp:equivalent}
\indent In this section, we provide an equivalent proof and illustrate the deep connection between the Normalized \textit{DC-RSBA} and proposed \textit{NW-RSBA} method. 
\subsection{Pre-definition}
\label{section:pre_definition}
\indent Recall the Eq.~(\ref{equation:proof3}), we define a new vector $\boldsymbol{\chi}_i^j$ for convenience:
\begin{equation}
\begin{aligned}
\boldsymbol{\chi}_i^j &= \frac{\partial \Pi (\mathbf{R}^j(r_i^j)\mathbf{P}_{i} + \mathbf{t}^j(r_i^j))}{\partial n_r} \\&= \frac{\partial \Pi (\mathbf{R}^j(r_i^j)\mathbf{P}_{i} + \mathbf{t}^j(r_i^j))}{\partial {r}_i^j},
\label{supp:b_chi_definition}
\end{aligned}
\end{equation}

\noindent then we can reformulate Eq.~(\ref{supp:sigma_c}) as:
\begin{equation}
\begin{aligned}
\mathbf{C}_i^j &= \begin{bmatrix}1 &0 \\ 0 &1\end{bmatrix} -\frac{\partial \Pi (\mathbf{R}^j(r_i^j)\mathbf{P}_{i} + \mathbf{t}^j(r_i^j))}{\partial n_r}\begin{bmatrix}0 &1 \end{bmatrix} \\&= \begin{bmatrix}1 & - {\boldsymbol{\chi}_i^j}(1) \\  0 &1 - {\boldsymbol{\chi}_i^j}(2)\end{bmatrix},
\label{supp:b_c_definition}
\end{aligned} 
\end{equation}  
where ${\boldsymbol{\chi}_i^j}(1) \ {\boldsymbol{\chi}_i^j}(2)$ are the first and second row of ${\boldsymbol{\chi}_i^j}$, and its inverse formulation is defined as:
\begin{align}
{\mathbf{C}_i^j}^{-1}  = \begin{bmatrix}1 & \frac{{\boldsymbol{\chi}_i^j}(1)}{1 - {\boldsymbol{\chi}_i^j}(2)} \\  0 &\frac{1}{1 - {\boldsymbol{\chi}_i^j}(2)}\end{bmatrix}.
\label{supp:b_inverse_c_definition}
\end{align}
\noindent Then we can define a new rectified image coordinate vector $\begin{bmatrix}{c_i^j}^" &{r_i^j}^"\end{bmatrix}^{\top}$ which represents the virtual image point after weighting.
\begin{align}
\begin{bmatrix}{c}_i^j\\{r}_i^j\end{bmatrix} - \begin{bmatrix}{c_i^j}^"\\{r_i^j}^"\end{bmatrix}= {\mathbf{C}_i^j}^{-1}(\begin{bmatrix}{c}_i^j\\{r}_i^j\end{bmatrix} - \begin{bmatrix}{c_i^j}^{'}\\{r_i^j}^{'}\end{bmatrix}),
\label{supp:equivalent_definition}
\end{align}
\noindent where ${\begin{bmatrix}{c_i^j}^{'}&{r_i^j}^{'}\end{bmatrix}}^\top$ is the projection image point with image measurement ${\begin{bmatrix}{c_i^j}&{r_i^j}\end{bmatrix}}^\top$, which is defined as:
\begin{align}
{\begin{bmatrix}{c_i^j}^{'}&{r_i^j}^{'}\end{bmatrix}}^\top = \Pi (\mathbf{R}^j(r_i^j)\mathbf{P}_{i} + \mathbf{t}^j(r_i^j)).
\label{supp:equivalent_supp_definition}
\end{align}
\noindent Our goal is to prove that using rectified coordinates as the observed image point will project on the same image point with the normalized measurement-based projection. We can summarize such equivalent as the following equation:
\begin{align}
\begin{bmatrix}
{c_i^j}^" & {r_i^j}^"
\end{bmatrix}^{\top} = \Pi (
\mathbf{R}^j({r_i^j}^")\mathbf{P}_{i} +  \mathbf{t}^j({r_i^j}^")
).
\label{supp:equivalent_goal}
\end{align}
We follow the schedule that firstly solves the Eq.~(\ref{supp:equivalent_definition}) to get the rectified image coordinate vector $\begin{bmatrix}{c_i^j}^" &{r_i^j}^"\end{bmatrix}^{\top}$, use the rectified image coordinate vector in normalized measurement based projection to get the projection point and finally check out whether it is the same.
\subsection{New Rectified Image Coordinate Solution}
\label{section:proof_r}
\noindent We solve Eq.~(\ref{supp:equivalent_definition}) consequently. Firstly, we solve ${r_i^j}^"$.
\begin{align}
 (1 - {\boldsymbol{\chi}_i^j}(2))({r}_i^j -{r_i^j}^")= ({r}_i^j - {r_i^j}^{'}),
\end{align}
\begin{equation}
\begin{aligned}
{r_i^j}^"&= {r_i^j}+\frac { {r_i^j}^{'}-{r}_i^j}{1 - {\boldsymbol{\chi}_i^j}(2)}\\&={{r}_i^j}^{'}+\frac {{\boldsymbol{\chi}_i^j}(2)( {r_i^j}^{'}-{r}_i^j) }{1 - {\boldsymbol{\chi}_i^j}(2)},
\end{aligned} 
\end{equation}  
We then substitute ${r_i^j}^"$ to solve ${c_i^j}^"$.
\begin{equation}
\begin{aligned}
({c}_i^j -{c_i^j}^") = \frac{{\boldsymbol{\chi}_i^j}(1)}{1 - {\boldsymbol{\chi}_i^j}(2)}({r}_i^j -{r_i^j}^{'}) +({c}_i^j - {c_i^j}^{'}),
\end{aligned} 
\end{equation}  
\begin{align}
{c_i^j}^"= {c_i^j}^{'} +  \frac{{\boldsymbol{\chi}_i^j}(1)({r_i^j}^{'} - {r}_i^j )}{1 - {\boldsymbol{\chi}_i^j}(2)}.
\end{align}

\subsection{Proof of equivalent after projection}
\label{section:proof_c}
\noindent We then substitute ${r_i^j}^"$ and ${c_i^j}^"$ in normalized measurement based projection.
\begin{equation}
\resizebox{0.88\linewidth}{!}{
	\begin{math}
\begin{aligned}
    \begin{bmatrix}
    {c_i^j}^{new}\\
    {r_i^j}^{new}
    \end{bmatrix} &=  \Pi (
\mathbf{R}^j({r_i^j}^")\mathbf{P}_{i} +  \mathbf{t}^j({r_i^j}^")
)\\&=\Pi (
\mathbf{R}^j({r_i^j} + \frac { {r_i^j}^{'}-{r}_i^j }{1 - {\boldsymbol{\chi}_i^j}(2)})\mathbf{P}_{i} +  \mathbf{t}^j({r_i^j} + \frac { {r_i^j}^{'}-{r}_i^j }{1 - {\boldsymbol{\chi}_i^j}(2)})
)   \\ &\approx\begin{bmatrix}{c_i^j}^{'}\\{r_i^j}^{'}\end{bmatrix}+(\frac{\partial  \Pi (
\mathbf{R}^j({r_i^j})\mathbf{P}_{i} +  \mathbf{t}^j({r_i^j}))}{\partial {r}_i^j})\frac { {r_i^j}^{'}-{r}_i^j }{1 - {\boldsymbol{\chi}_i^j}(2)}\\&= \begin{bmatrix}{c_i^j}^{'}\\{r_i^j}^{'}\end{bmatrix}+ {\boldsymbol{\chi}_i^j} \frac {{r_i^j}^{'}-{r}_i^j}{1 - {\boldsymbol{\chi}_i^j}(2)}\\&=\begin{bmatrix}{c_i^j}^"\\{r_i^j}^"\end{bmatrix}
\end{aligned} 
\end{math}
}
\label{supp:prove}
\end{equation}

\subsection{Connection between Normalized DC-
RSBA and NW-RSBA}
\label{section:connection}
\indent From Eq.~(\ref{supp:prove}), we can get a such summary
 that Normalized \textit{DC-RSBA} is equivalent to the proposed \textit{NW-RSBA} mathematical. It is amazing to view that although the two formulations are totally different from each other, they both bring in the implicit rolling shutter constraint to optimization. However, although these two methods are equivalent to each other, our proposed \textit{NW-RSBA} is much easier and faster to solve since we provide detailed analytical Jacobian matrices. 
 \\
 \\
\section{Proposed NW-RSBA Algorithm Pipeline }
In this section, we provide a detailed bundle adjustment algorithm  pipeline with the standard Gauss-Newton least square solver.
\listofalgorithms
\begin{algorithm}
	\SetKwInOut{Input}{Input}
	\SetKwInOut{Output}{Output}
	
	\KwIn{Initial rolling shutter camera poses $\{\mathbf{R}^1, \mathbf{t}^1, \boldsymbol{\omega}^1, \mathbf{d}^1\}$,...,$\{\mathbf{R}^j, \mathbf{t}^j, \boldsymbol{\omega}^j, \mathbf{d}^j\}$, points $\mathbf{P}_1$,...,$\mathbf{P}_i$ as $\boldsymbol{\theta}$ and point measurement in normalized image coordinate $\mathbf{q}^{1...j}_{1...i}$}
	\KwOut{Refined parameters $\boldsymbol{\theta}^{*}$}
	\While{(not reach max iteration) and (not satisfy stopping criteria)}
	{

		\For{ Each camera $j\in \mathcal{F}$}
		{
			\For{ Each point $i\in \mathcal{P}_j$}
			{
				Calculate weighted reprojection error $\mathbf{\hat{e}}^j_i $ using Alg.~\ref{algorithm:NW_error}\;
				Construct Jacobian matrix $\mathbf{J}^j_i$ using Alg.~\ref{algorithm:J_matrix}\;
				Parallel connect $\mathbf{J}^j_i$ to $\mathbf{J}$\;
				Stack $\mathbf{\hat{e}}^{j}_{i}$ into $\mathbf{\hat {e}}$\;
			}
			
		}
		Solve normal equation $\mathbf{J}^{\top}\mathbf{J} \boldsymbol{\delta} =-\mathbf{J}^{\top}\mathbf{\hat {e}}$ using Alg.~\ref{algorithm:Series_connection_J}\;
		Update camera poses and points parameters $\boldsymbol{\theta}$ using $\boldsymbol{\delta}$\;
	}	
	\caption{Normalized Weighted RSBA}
	\label{algorithm:RSBA_pipeline}
\end{algorithm}
\begin{algorithm}[t]
	\SetKwInOut{Input}{Input}
	\SetKwInOut{Output}{Output}
	
	\Input{Rolling shutter camera poses $\{\mathbf{R}^j, \mathbf{t}^j, \boldsymbol{\omega}^j, \mathbf{d}^j\}$, points $\mathbf{P}_i$ and point measurement in normalized image coordinate $\mathbf{q}^j_i$}
	\Output{Normalized weighted error $\mathbf{\hat{e}}^j_i $}
	Compute weight matrix $\mathbf{C}^j_i $ using Eq.~(\ref{supp:sigma_c})\;
	Compute standard reprojection error $\mathbf{e}^j_i $\;
	Return normalized weighted reprojection error $\mathbf{\hat{e}}^j_i $ using Eq.~(\ref{supp:supp_J_matrix})\;
	
	\caption{Computation of weighted reprojection error}
	\label{algorithm:NW_error}
\end{algorithm}
\begin{algorithm}[t]
	\SetKwInOut{Input}{Input}
	\SetKwInOut{Output}{Output}
		\Input{RS camera poses $\{\mathbf{R}^j, \mathbf{t}^j, \boldsymbol{\omega}^j, \mathbf{d}^j\}$, points coordinate  $\mathbf{P}_i$ and point measurement in normalized image coordinate $\mathbf{q}^j_i$}
	\Output{Jacobian matrix $\mathbf{J}^j_i$}
	
	Calculate $\frac {\partial{\hat{\mathbf{e}}_i^j}}{\partial\mathbf{P}_{i}}$ using Eq.~(\ref{supp:PD_pi})\;

	Calculate $\frac {\partial{\hat{\mathbf{e}}_i^j}}{\partial{\boldsymbol\xi}^{j}}$  using Eq.~(\ref{supp:PD_ri})\;

	Calculate $\frac {\partial{\hat{\mathbf{e}}_i^j}}{\partial\mathbf{t}^{j}}$ using Eq.~(\ref{supp:PD_ti})\;
	
	Calculate $\frac {\partial{\hat{\mathbf{e}}_i^j}}{\partial\boldsymbol{\omega}^{j}}$ using Eq.~(\ref{supp:PD_wi})\;
	
	Calculate $\frac {\partial{\hat{\mathbf{e}}_i^j}}{\partial\mathbf{d}^{j}}$ using Eq.~(\ref{supp:PD_di})\;
	
	Construct
	\begin{equation*}
		\begin{aligned}
		\mathbf{J}^j_i &= \begin{bmatrix} \mathbf{J}_{i,rs}^j & \mathbf{J}_{i,gs}^j & \mathbf{J}_{i,p}^j \end{bmatrix}  \\
		&= \begin{bmatrix}
			[\frac {\partial{\hat{\mathbf{e}}_i^j}}{\partial{\boldsymbol{\omega}}^{j}},\frac {\partial{\hat{\mathbf{e}}_i^j}}{\partial{\mathbf{d}}^{j}}] & [\frac {\partial{\hat{\mathbf{e}}_i^j}}{\partial{\boldsymbol{\xi}}^{j}},\frac {\partial{\hat{\mathbf{e}}_i^j}}{\partial{\mathbf{t}}^{j}}] & \frac {\partial{\hat{\mathbf{e}}_i^j}}{\partial{\mathbf{P}}_{i}}
		\end{bmatrix};
		\end{aligned}
	\end{equation*}
	
	\caption{Computation of Jacobian matrix}
	\label{algorithm:J_matrix}
\end{algorithm}

\begin{algorithm}[t]
	\SetKwInOut{Input}{Input}
	\SetKwInOut{Output}{Output}
	
	\Input{Jacobian matrix $\mathbf{J}$ and weighted error vector $\mathbf{\hat{e}}$}
	\Output{Updated vector $\boldsymbol{\delta}$}
	Compute Schur complement matrix $\textbf{S}_p$ and $\textbf{S}_{rs}$\;
	Compute auxiliary vectors $\textbf{t}^{*}$ and $\textbf{u}^{*}$\;
	
	Solve normal equation cascadingly:
	\begin{itemize}
		\item Get $\boldsymbol{\delta}_{rs}$ by solving $\mathbf{S}_{rs}\boldsymbol{\delta}_{rs} = -\mathbf{t}^{*}$\;
		\item Get $\boldsymbol{\delta}_{gs}$ by solving $\mathbf{U}^{*} \boldsymbol{\delta}_{gs}= - \textbf{u}^{*} - \mathbf{S}^{*\top}\boldsymbol{\delta}_{rs}$\;
		\item Get $\boldsymbol{\delta}_{p}$ by solving $\mathbf{V} \boldsymbol{\delta}_{p}= - \mathbf{v} - \mathbf{T}^{\top}\boldsymbol{\delta}_{rs} - \mathbf{W}^{\top} \boldsymbol{\delta}_{gs}$\;
		
		\item Stack $\boldsymbol{\delta}_{gs}$ $\boldsymbol{\delta}_{rs}$ $\boldsymbol{\delta}_{p}$ into $\boldsymbol{\delta}$\;
	\end{itemize}

	\caption{Solve the normal equation using two-stage Schur complement}
	\label{algorithm:Series_connection_J}
\end{algorithm}

\section{Experimental Settings and Evaluation Metrics}
\label{supp:Experimental_setting}
In this section, we provide detailed experiment settings and evaluation metrics used in synthetic data experiments and real data experiments.

\subsection{Synthetic Data}
\label{section：Synthetic_Data}

\noindent \textbf{Experimental Settings.} We simulate $5$ RS cameras located randomly on a sphere with
a radius of 20 units pointing to a cubical scene with 56 points. The RS image size is 1280 × 1080 px, the focal length is about 1000, and the optical center is at the center of the image domain. We compare all methods by varying the speed, the noise on image measurements, and the readout direction. The results are obtained after collecting the errors over 300 trials each epoch. The default setting is 10 deg/frame and 1 unit/frame for angular and linear velocity, standard covariance noise.

\begin{itemize}
	
	\item \textbf{{Varying Speed}}: We evaluate the accuracy of five approaches against increasing angular and linear velocity from 0 to 20 deg/frame and 0 to 2 units/frame gradually, with random directions.
	
	\item \textbf{{Varying Noise Level}}: We evaluate the accuracy of five approaches against increasing noise level from 0 to 2 pixels.
	
	\item \textbf{{Varying Readout Direction}}: We evaluate the robustness of five methods with an RS critical configuration. Namely, the readout directions of all views are almost parallel. Thus, we vary the readout directions of the cameras from parallel to perpendicular by increasing the angle from 0 to 90 degrees.
	
	\item \textbf{{Runtime}}: We compare the time cost of all methods against increasing the number of cameras from 50 - 250 with a fixed number of points.

\end{itemize}

\noindent \textbf{Evaluation metrics.} In this section, we use three metrics to evaluate the performances, namely reconstruction error, rotation error, and translation error.

\begin{itemize}
	
	\item \textbf{{Reconstruction Error $\mathbf{e}_{\text{point}}$}}: We use the reconstruction error to measure the difference between computed and ground truth 3D points, which is defined as:\\
	$\mathbf{e}_{\text{point}} = {\begin{Vmatrix}
\mathbf{P} - \mathbf{P}_{\text{GT}}
\end{Vmatrix}}^2$. 
	
	\item \textbf{{Rotation Error  $\mathbf{e}_{\text{rot}}$}}: We utilize the geodesic distance to measure the error between optimized rotation and ground truth. The error is defined as:\\ $\mathbf{e}_{\text{rot}} = \arccos((\text{tr}(\mathbf{R}\mathbf{R}_{\text{GT}}^{\top})-1)/2)$.
	
	\item \textbf{{Translation Error $\mathbf{e}_{\text{trans}}$}}: We use normalized inner product distance to measure the error between optimized translation and ground truth, which is defined as:\\ $\mathbf{e}_{\text{trans}} = \arccos(\mathbf{t}^{\top}\mathbf{t}_{\text{GT}}/(\left \| \mathbf{t} \right \| \left \| \mathbf{t}_{\text{GT}} \right \|))$.

\end{itemize}

\subsection{Real Data}

\noindent \textbf{Datasets Settings.} We compare all the RSC methods in the following publicly available RS datasets.

\begin{itemize}
	
	\item \textbf{\textit{WHU-RSVI}}: WHU dataset	\footnote{\href{http://aric.whu.edu.cn/caolike/2019/11/05/the-whu-rsvi-dataset/}{http://aric.whu.edu.cn/caolike/2019/11/05/the-whu-rsvi-dataset/}} was published in~\cite{cao2020whu} and provided ground truth synthetic GS images, RS images and camera poses.  
	
	\item \textbf{\textit{TUM-RSVI}}: The TUM RS dataset\footnote{\href{https://vision.in.tum.de/data/datasets/rolling-shutter-dataset}{https://vision.in.tum.de/data/datasets/rolling-shutter-dataset}} was published in~\cite{schubert2019rolling} and contained time-synchronized global-shutter, and rolling-shutter images captured by a non-perspective camera rig and ground-truth poses recorded by motion capture system for ten RS video sequences. 
	
\end{itemize}

\noindent \textbf{Evaluation metrics.} In this section, we use three metrics to evaluate the performances, namely absolute trajectory error, tracking duration and real-time factor.  

\begin{itemize}
	\item \textbf{Absolute trajectory error (ATE).} We use the absolute trajectory error (\textit{\textbf{ATE}})~\cite{schubert2019rolling} to evaluate the VO results quantitatively. Given ground truth frame positions $\bar{\mathbf{c}}_i\in \mathbb{R}^{3}$ and corresponding Orb-SLAM~\cite{mur2015orb} tracking results $\mathbf{c}_i\in \mathbb{R}^{3}$ using corrected sequence by each RSC method. It is defined as
	\begin{equation}\label{equation:VO_metric}
	e_{\text{ate}} = \min_{\mathbf{T}\in \text{Sim}(3)}\sqrt{\frac{1}{n}\sum_{i=1}^{n} \left \| \mathbf{T}(\textbf{c}_i) -  \bar{\textbf{c}}_i\right \|},
	\end{equation} 
	where $\mathbf{T}\in \text{Sim}(3)$ is a similarity transformation that aligns the estimated trajectory with the ground truth one since the scale
	is not observable for monocular methods. We run each method 20 times on each sequence to obtain the ATE $e_{\text{ate}}$.
	
	\item \textbf{Tracking duration (DUR).} Besides, we find out that some RSC solutions provide the results of corrections that are even worse than the original input RS frames. This leads to failure in tracking and makes Orb-SLAM interrupt before the latest capture frame. Therefore, we use the ratio of the successfully tracked frames out of the total frames $\textit{\textbf{DUR}}$ as an evaluation metric. 
	
	\item \textbf{Realtime factor $\epsilon$.} The realtime factor $\epsilon$ is calculated as the sequence's actual duration divided by the algorithm's processing time. 
\end{itemize}

\begin{figure*}[t]
	\begin{center}
		\includegraphics[width=.9\linewidth]{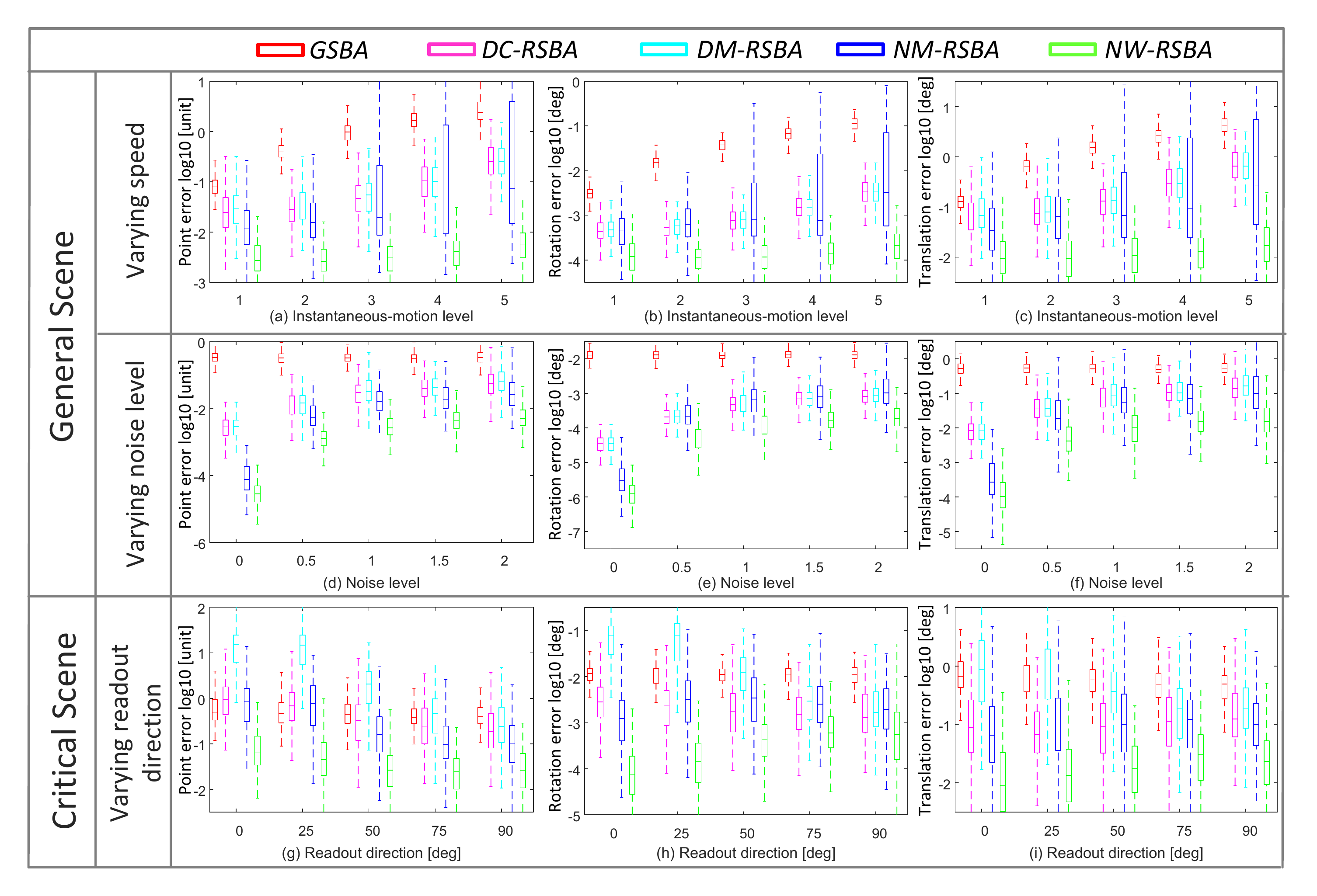}
	\end{center}
	\caption{Camera pose ($2^\text{nd}$ and $3^\text{rd}$ columns) and reconstruction ($1^\text{st}$ column) errors of \textit{GSBA}, \textit{DC-RSBA}, \textit{DM-RSBA}, \textit{NM-RSBA} and \textit{NW-RSBA} with increasing angular and linear velocity  ($1^\text{st}$ row) and noise levels in the image ($2^\text{nd}$ row) in general scenes, also with increasing readout directions in degeneracy scene ($3^\text{rd}$ row).}
	\label{fig:synthetic1}
\end{figure*} 

\begin{figure*}[t]
	\begin{center}
		\includegraphics[width=.95\linewidth]{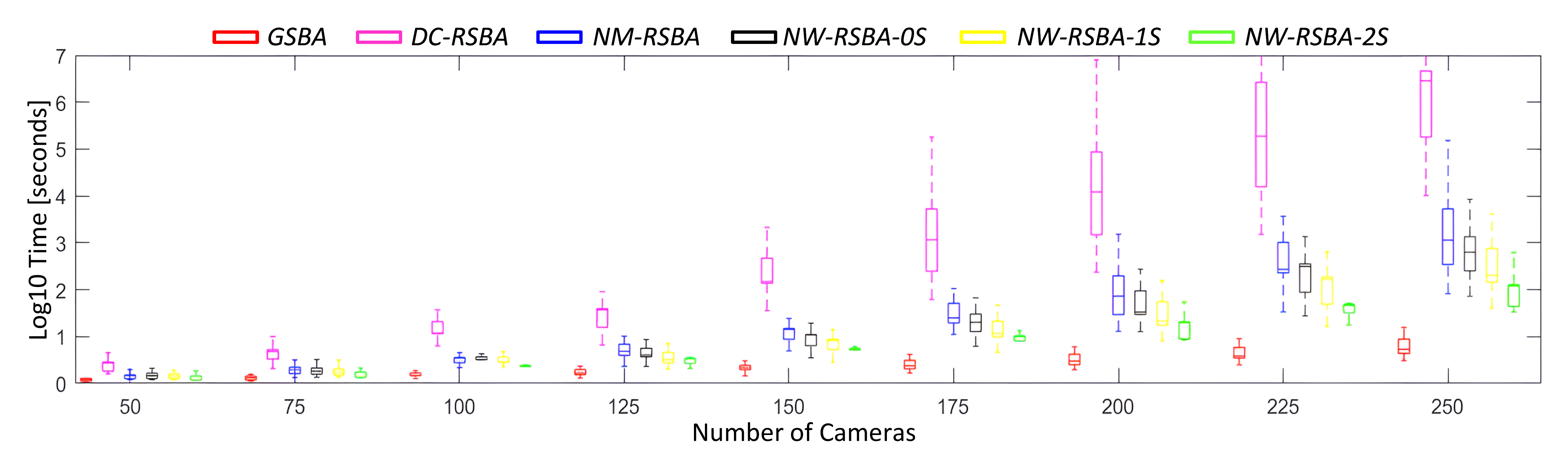}
	\end{center}
	\caption{Time cost of \textit{GSBA}~\cite{lourakis2009sba}, \textit{DC-RSBA}~\cite{lao2018robustified}, \textit{NM-RSBA}~\cite{Albl2016}, \textit{NW-RSBA-0S} (without Schur complement), \textit{NW-RSBA-1S} (one-stage Schur complement to Jacobian matrices with series connection), and proposed \textit{NW-RSBA-2S} (two-stage Schur complement to Jacobian matrices with parallel connection) with increasing camera number.}
	\label{fig:time}
\end{figure*} 
\begin{figure*}[t]
	\begin{center}
		\includegraphics[width=.9\linewidth]{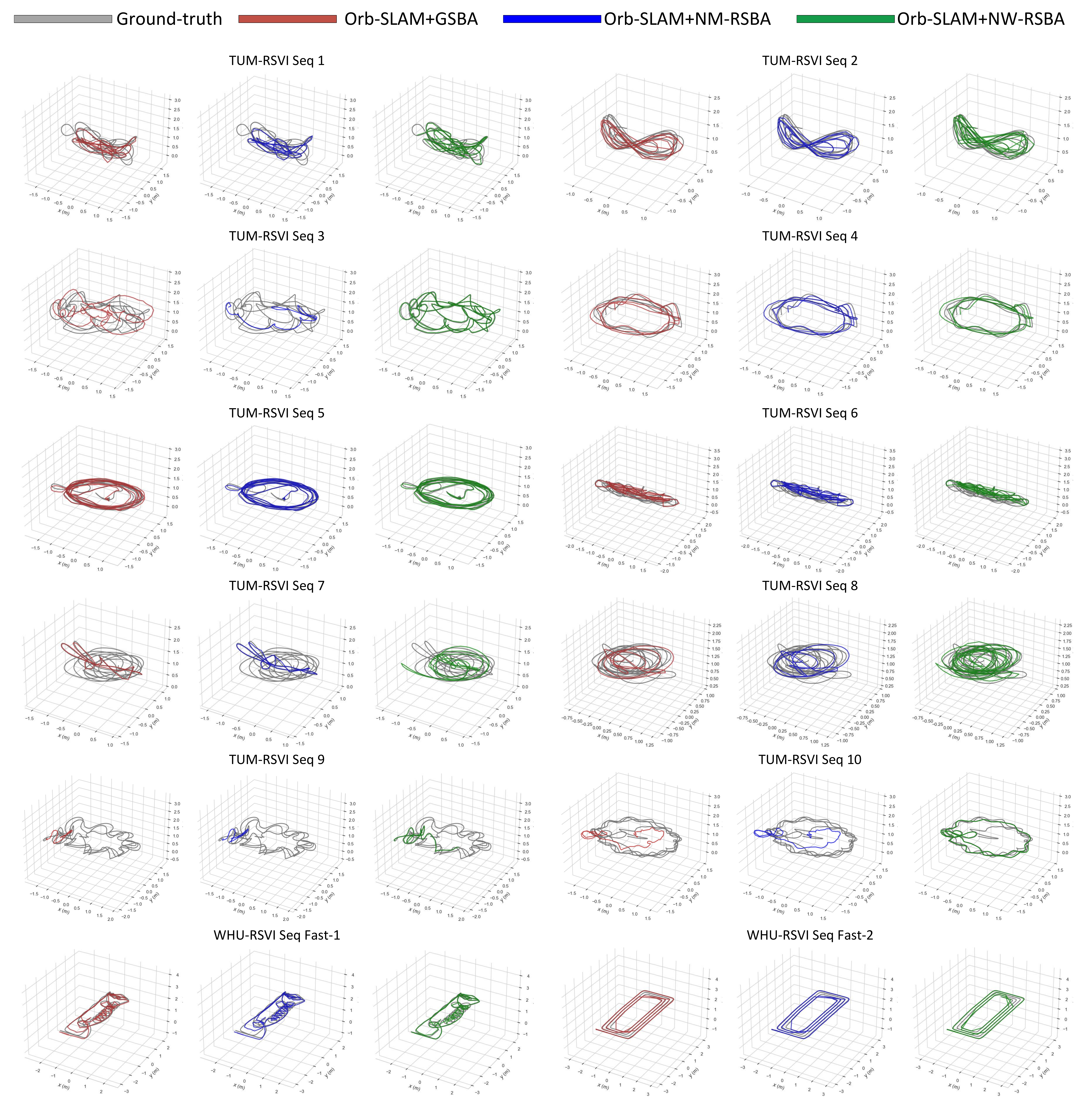}
	\end{center}
	\caption{Ground truth and trajectories estimated by \textit{GSBA}~\cite{lourakis2009sba}, \textit{NM-RSBA}~\cite{Albl2016} and proposed \textit{NW-RSBA} after Sim(3) alignment on 10 sequences from TUM-RSVI~\cite{schubert2019rolling} and 2 sequences from WHU-RSVI~\cite{cao2020whu} datasets.}
	\label{fig:real_slam_traj}
\end{figure*} 

\end{document}